\documentclass[10pt,twocolumn,letterpaper]{article}

\usepackage{amsfonts}
\usepackage{cvpr}
\usepackage{times}
\usepackage{epsfig}
\usepackage{subfigure}
\usepackage{graphicx}
\usepackage{amsmath}
\usepackage{amssymb}
\usepackage{algorithm}
\usepackage{algpseudocode}
\usepackage{multirow}
\usepackage[pagebackref=true,breaklinks=true,letterpaper=true,colorlinks,bookmarks=false]{hyperref}

\cvprfinalcopy % *** Uncomment this line for the final submission

\def\cvprPaperID{3867} % *** Enter the CVPR Paper ID here

\ifcvprfinal\fi
\begin{document}

%%%%%%%%% TITLE
\title{Awesome Typography: Statistics-Based Text Effects Transfer}

\author{Shuai Yang, Jiaying Liu, Zhouhui Lian and Zongming Guo\\
Institute of Computer Science and Technology, Peking University, Beijing, China\\
}

\maketitle
%\thispagestyle{empty}

%%%%%%%%% ABSTRACT
\begin{abstract}
\vspace{-1mm}
In this work, we explore the problem of generating fantastic special-effects for the typography.
It is quite challenging due to the model diversities to illustrate varied text effects for different characters.
To address this issue, our key idea is to exploit the analytics on the high regularity of the spatial distribution for text effects to guide the synthesis process.
Specifically, we characterize the stylized patches by their normalized positions and the optimal scales to depict their style elements. Our method first estimates these two features and derives their correlation statistically.
They are then converted into soft constraints for texture transfer to accomplish adaptive multi-scale texture synthesis and to make style element distribution uniform.
It allows our algorithm to produce artistic typography that fits for both local texture patterns and the global spatial distribution in the example.
Experimental results demonstrate the superiority of our method for various text effects over conventional style transfer methods. In addition, we validate the effectiveness of our algorithm with extensive artistic typography library generation.
\vspace{-4mm}
\end{abstract}

%%%%%%%%% BODY TEXT

\section{Introduction}
\begin{figure}
  \centering
    \subfigure{
    \includegraphics[width=0.99\linewidth]{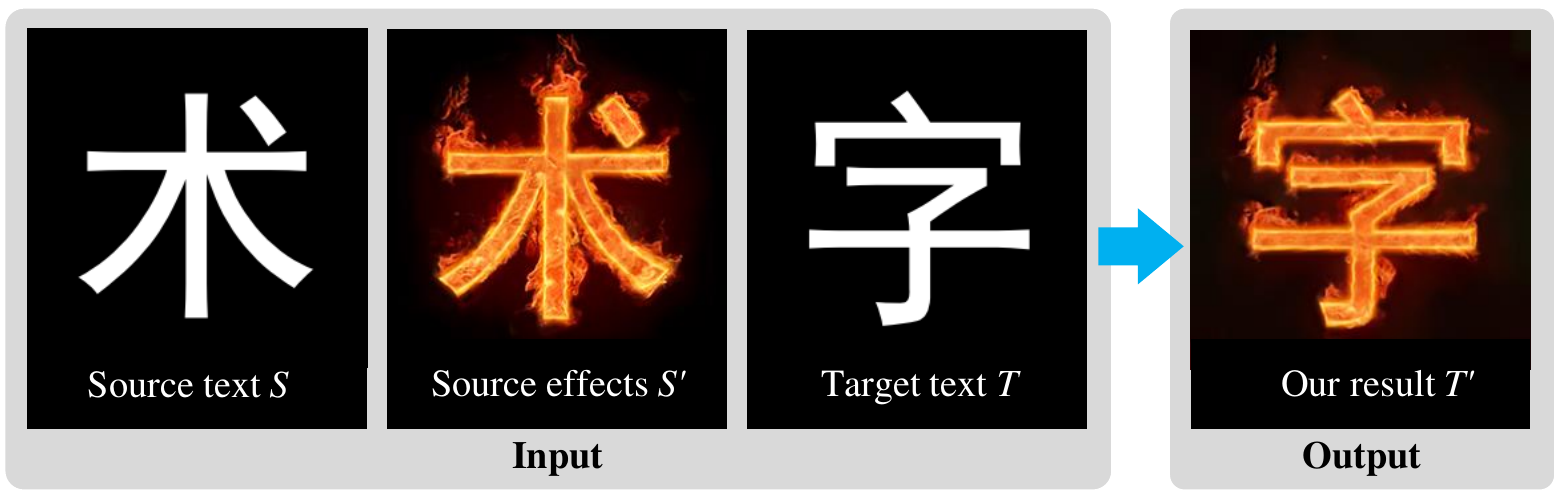}}\vspace{-3mm}
    \subfigure{
    \includegraphics[width=0.98\linewidth]{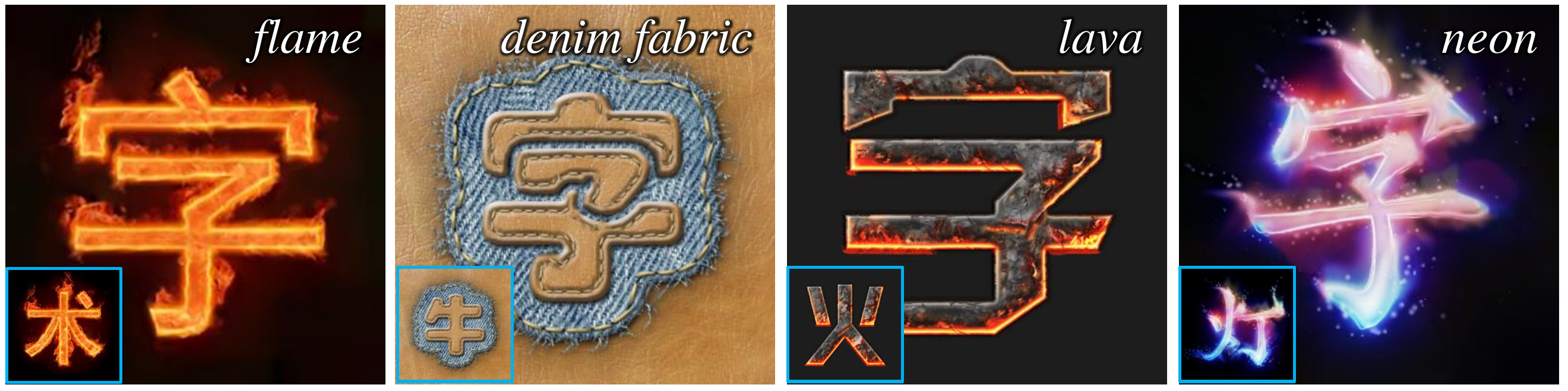}}
  \caption{Overview: Our method takes as input the source text image $S$, its counterpart stylized image $S'$ and the target text image $T$, then automatically generates the target stylized image $T'$ with the special effects as in $S'$.}\label{fig:overview}
  \vspace{-6mm}
\end{figure}

% 背景
Typography is the technology to design the special text effects to render the character into an original and unique artwork.
These amazing text styles include basic effects such as \textit{shadows}, \textit{outlines}, \textit{colors} and sophisticated effects such as burning \textit{flames}, flowing \textit{smokes}, multicolored \textit{neons}, as shown in Fig.~\ref{fig:overview}. Texts decorated by well-designed special effects become much more attractive. It can also better reflect the thoughts and emotions from the designer. The beauty and elegance of text effects are well appreciated, making it widely used in the publishing and advertisement. However, creating vivid text effects requires a series of subtle processes by an experienced designer using the editing software: determine color styles, warp textures to match text shapes and adjust the transparency for visual plausibleness, \textit{etc}. These advanced editing skills are far beyond the abilities of most casual users. This practical requirement motivates our work: We investigate an approach to automatically transfer various fantastic text effects designed by artists onto raw plain texts, as shown in Fig.~\ref{fig:overview}.

% 相关工作
Text effects transfer is a brand new sub-topic of style transfer. Style transfer can be related to color and texture transfer, respectively. Color transfer matches global \cite{Reinhard2001Color} or local \cite{Tai2005Local} color distributions of the target and source images. Texture transfer relies on texture synthesis technologies, where the texture generation is constrained by guidance images. Meanwhile, texture synthesis can be divided into two categories: non-parametric methods \cite{efros1999texture,Efros2001Image,Kwatra2003Graphcut,Space-time} and parametric methods \cite{Julesz1983Textons,gatys2015texture,Li2016Combining,Goodfellow2014Generative}. The former generates new textures by resampling pixels or patches from the original texture, while the latter models textures using statistical measurements and produces a new texture that shares the same parametric results with the original one.

% 问题
From a technical perspective, it is quite challenging and impractical to directly exploit the traditional style transfer methods to generate new text effects. The challenges lie in three aspects:
(i) The extreme diversity of the text effects and character shapes: The style diversity makes the transfer task difficult to model uniformly. Further, the algorithm should be robust to the tremendous variety of characters.
(ii) The complicated composition of style elements: A text effects image often contains multiple intertwined style elements (we call them \textit{text sub-effects}) that have very different textures and structures (see \textit{denim fabric} example in Fig.~\ref{fig:overview}) and need specialized treatments.
(iii) The simpleness of guidance images: The raw plain text as guidance gives few hints on how to place different sub-effects. Textures in the white text and black background regions may not hold the stationarity. This makes the traditional non-parametric texture-by-numbers method \cite{Hertzmann2001Image} fail, which has assumed textures to be stationary in each region of the guidance map. Meanwhile, the plain text image provides little semantic information. This makes the recent successful parametric deep-based style transfer methods \cite{gatys2016image,Li2016Combining} lose their advantages of representing high-level semantic information. For these reasons, conventional style transfer methods for general styles perform poorly on text effects.

% 我们的算法
In this paper, we propose a novel text effects transfer algorithm to address these challenges. The key idea is to analyze and model the distance-based essential characteristics of high-quality text effects and to leverage them to guide the synthesis process. The characteristics are summarized based on the analytics over dozens of well-designed text effects into a general prior. This prior guides our style transfer process to synthesize different sub-effects adaptively and to simulate their spatial distribution. All measurements are carefully designed to achieve certain robustness to the character shape. In addition, we further consider the psycho-visual factor to enhance image naturalness. In summary, our contributions are threefold:
\begin{itemize}\vspace{-0.5mm}
  \item We raise a brand new topic of text effects transfer that turns plain texts into fantastic artworks, which enjoys wide application scenarios such as picture creation on social networks and commercial graphic design. \vspace{-1.5mm}
  \item We perform investigation and analysis on well-designed typography and summarize the key distance-based characteristics for high-quality text effects. We model these characteristics mathematically to form a general prior that can be used to significantly improve the style transfer process for texts. \vspace{-1.5mm}
  \item We propose the first method to generate compelling text effects, which share both similar local texture patterns and the global spatial distribution with the source example, while preserving image naturalness.
\end{itemize}

\section{Related Work}

% 参考CVPR split and match / EGSR Unifying Color and Texture Transfer
\textbf{Color Transfer.} Pioneering color transfer methods \cite{Reinhard2001Color,Piti2007Automated} transfer color between images by matching their global color distributions. Subsequently, local color transfer is achieved based on segmentation \cite{Tai2005Local,Tai2007Soft} or user interaction \cite{Welsh2002Transferring} and it is further improved  using fine-grained patch \cite{Shih2013Data} or pixel \cite{Shih2014Style,park2016efficient} correspondences. Recently, color transfer \cite{Yan2016Automatic} and colorization \cite{Larsson2016Learning,zhang2016colorful} using deep neural networks have drawn people's attentions.

\textbf{Non-Parametric Texture Synthesis and Transfer.} Efros and Lueng \cite{efros1999texture} proposed a pioneering pixel-by-pixel synthesis approach based on sampling similar patches. The subsequent works improve it in quality and speed by synthesizing patches rather than pixels. To handle the overlapped regions of neighboring patches for seamlessness, Liang \textit{et al}.~\cite{Liang2001Real} proposed to blend patches, and Efros and Freeman \cite{Efros2001Image} used dynamic programming to find an optimal separatrix in overlapped regions, which is further improved via graph cut \cite{Kwatra2003Graphcut}. Unlike previous methods that synthesize textures in a local manner, recent techniques synthesize globally using objective functions. A coherence-based function \cite{Space-time} is proposed to synthesize textures in an iterative coarse-to-fine fashion. This method performs patch matching and voting operations alternately and achieves good local structures. It is
%then accelerated using PatchMatch algorithm \cite{PatchMatch} and
then extended to adapt to non-stationary textures through patch geometric and photometric transformations \cite{GeneralizedPatchMatch,ImageMelding}.

Texture transfer, also known as Image Analogies \cite{Hertzmann2001Image}, generates textures but also keeps the structure of the target image. %Given guidance maps,
Structures are usually preserved by reducing the differences between the source and target guidance maps \cite{Hertzmann2001Image,Okura2015Unifying}. In \cite{Luk2013Painting}, texture boundaries are synthesized in priority to constrain the structure. Frigo \textit{et al}.~\cite{Frigo2016Split} proposed an adaptive patch partition to precisely capture source textures and preserve target structures, followed by a Markov Random Field (MRF) function for global texture synthesis.

\begin{figure*}[t]
  \centering
    \subfigure[\textit{flame}]{
    \includegraphics[width=0.18\linewidth]{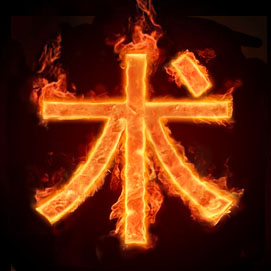}}
    \subfigure[Pattern distribution]{
    \includegraphics[width=0.18\linewidth]{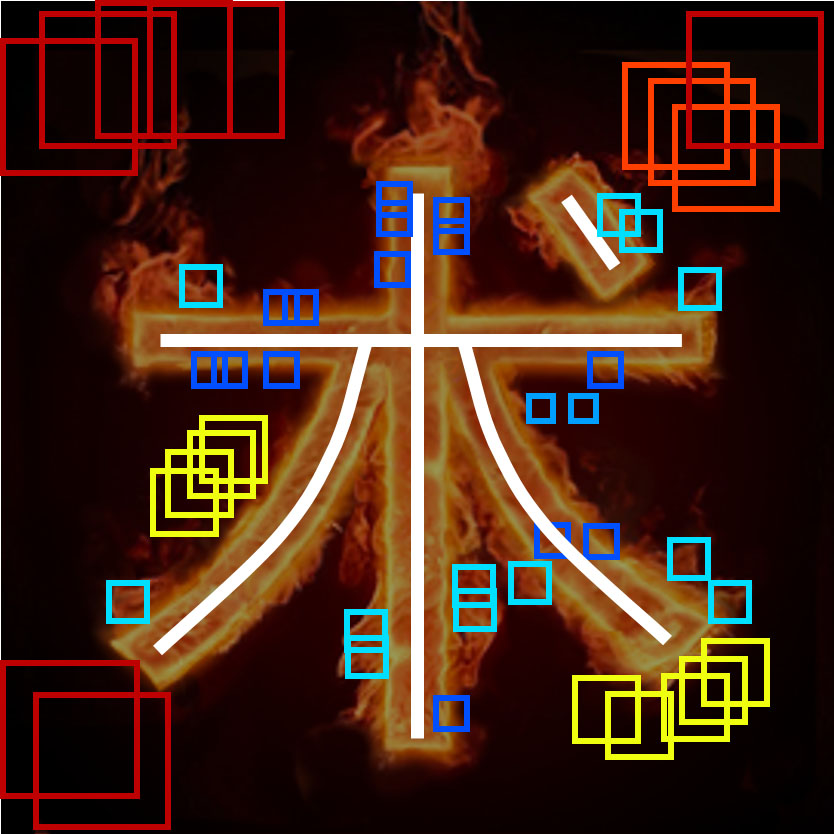}}
    \subfigure[\textit{denim fabric}]{
    \includegraphics[width=0.18\linewidth]{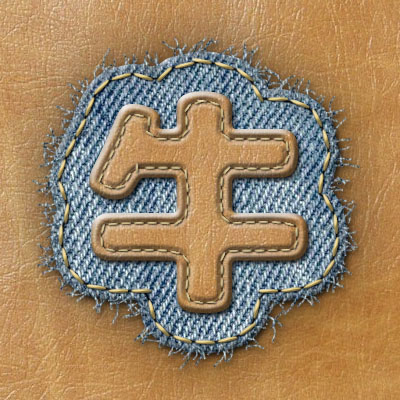}}
    \subfigure[Pattern distribution]{
    \includegraphics[width=0.18\linewidth]{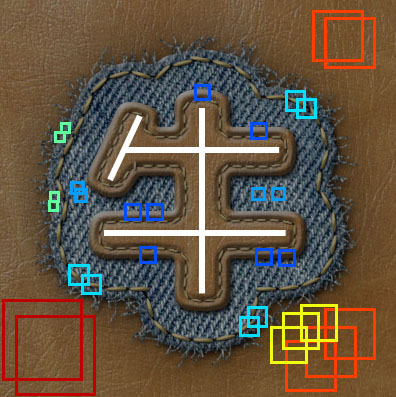}}
    \subfigure[Partition modes]{
    \includegraphics[width=0.18\linewidth]{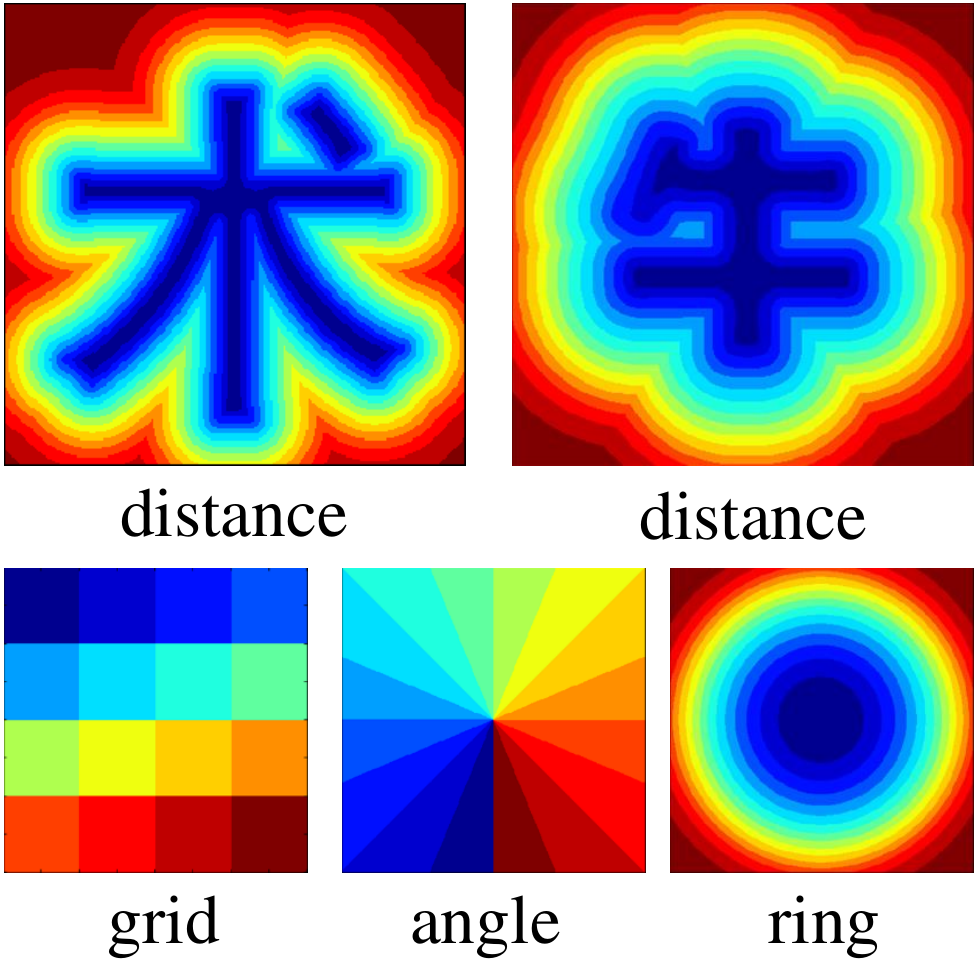}}\vspace{-2mm}
    \subfigure[Pixels in RGB space]{
    \includegraphics[width=0.24\linewidth]{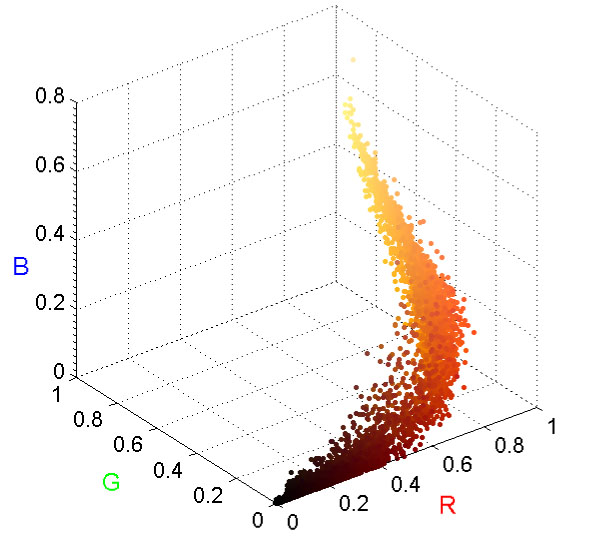}}
    \subfigure[Pixel distance distribution]{
    \includegraphics[width=0.24\linewidth]{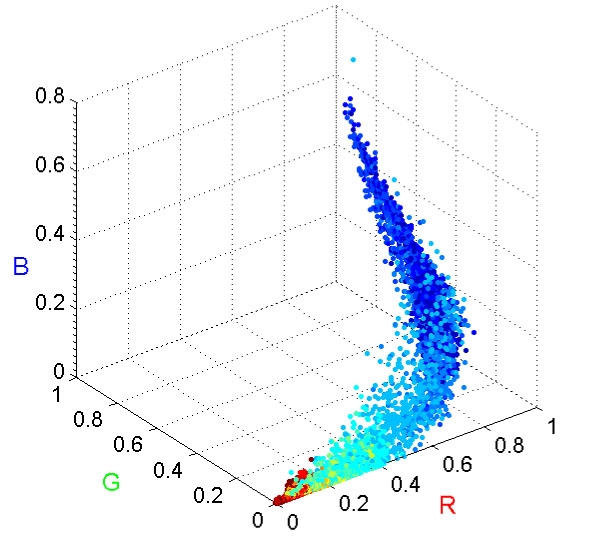}}
    \subfigure[Response curves (distance mode)]{
    \includegraphics[width=0.24\linewidth]{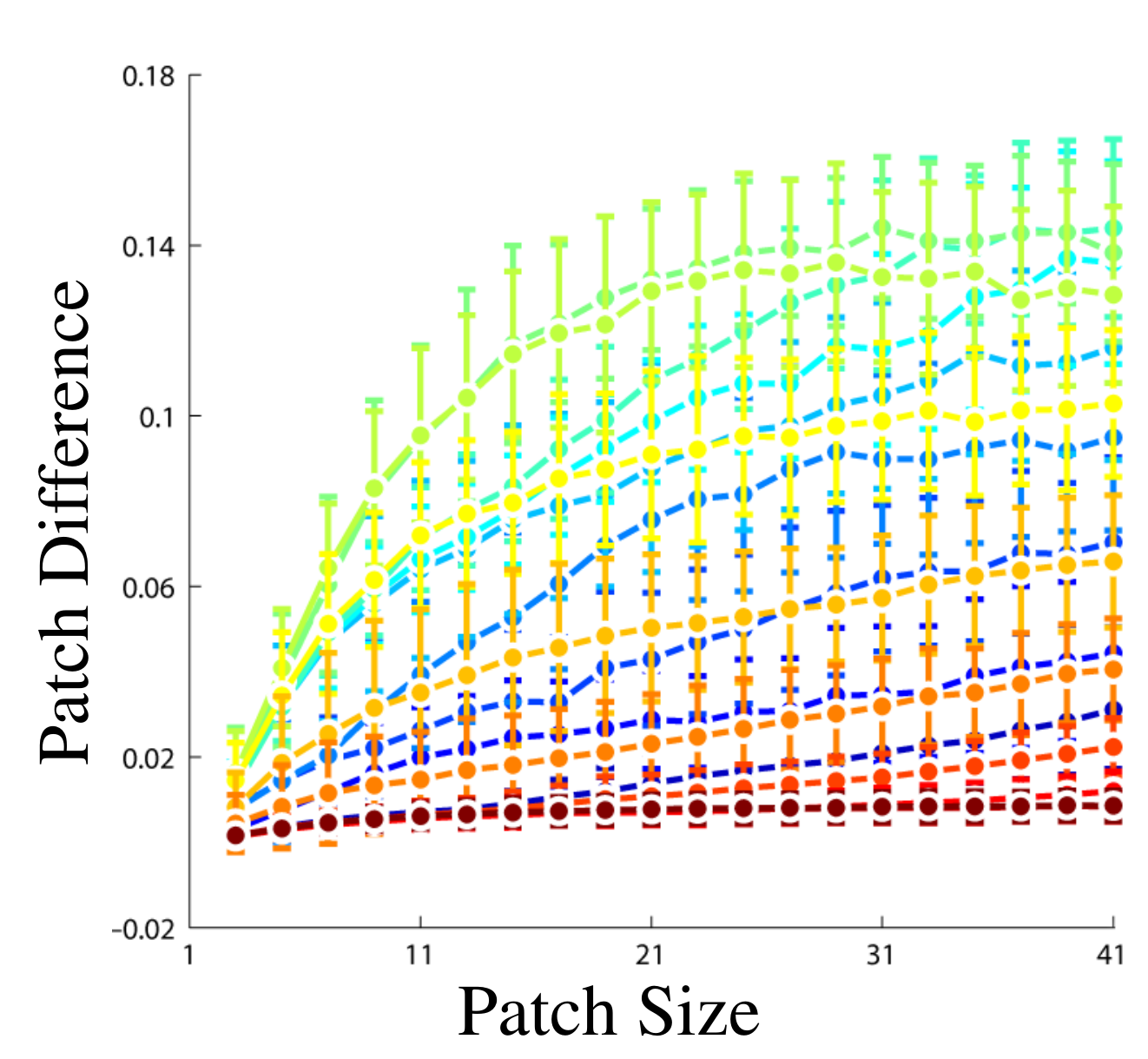}}
    \subfigure[Response curves (random mode)]{
    \includegraphics[width=0.24\linewidth]{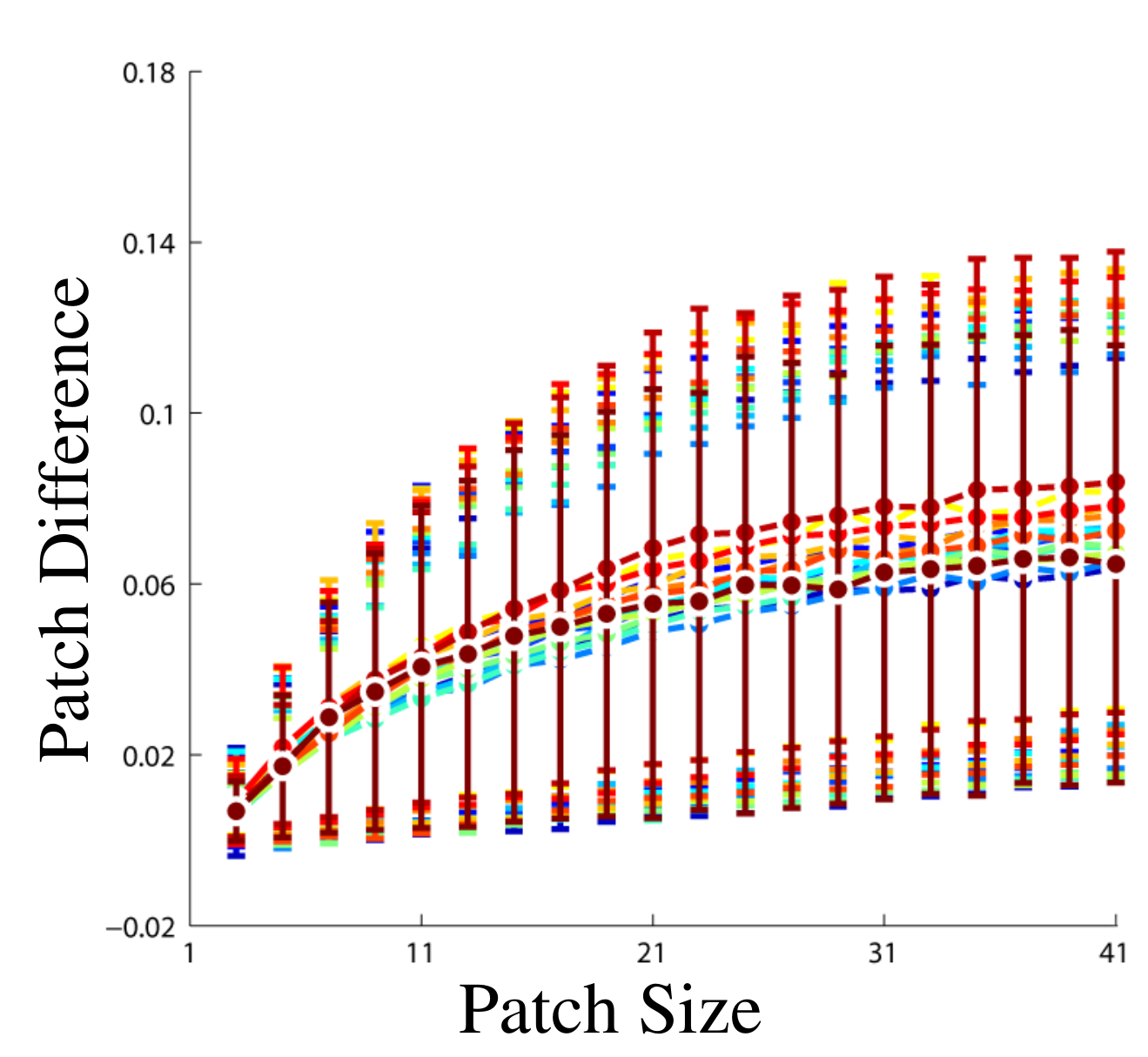}}
  \caption{Statistics of the text effects images. (a)(c) \textit{flame} and \textit{denim fabric} text effects. (b)(d) Textures with similar distances to the text skeleton (in white) trend to have similar patterns. (e) Pixels are divided into $N=16$ classes using different partition modes. (f)-(g) High correlation between pixel colors and distances: Pixels are distinguished from each other by their distances in RGB space. (h)-(i) High correlation between patch scales and distances: Patches with similar distances have uniform responses to changes of their size.}
  \label{fig:analysis}\vspace{-2mm}
\end{figure*}

\textbf{Parametric Texture Synthesis and Transfer.} The idea of modelling textures using statistical measurements has led to the development of textons and its variants \cite{Julesz1983Textons,Portilla2000A}. Nowadays, deep-based texture synthesis \cite{gatys2015texture} starts trending due to the great descriptive ability of deep neural networks. Gatys \textit{et al}.~proposed to use Gram-matrix in the Convolutional Neural Networks (CNNs) feature space to represent textures \cite{gatys2015texture} and adapt it to style transfer by incorporating content similarities \cite{gatys2016image}. This work presented the remarkable generic painting transfer technique and attracted many follow-ups in loss function improvement \cite{lin2015visualizing,Selim2016Painting} and algorithm acceleration \cite{Johnson2016Perceptual,ulyanov2016texture}. Recently, methods that replace the Gram-matrix by MRF regularizer is proposed for photographic synthesis \cite{Li2016Combining} and semantic texture transfer \cite{Champandard2016Semantic}. Meanwhile, Generative Adversarial Networks (GANs) \cite{Goodfellow2014Generative} provide another idea for texture generation by using discriminator and generator networks, which iteratively improve the model by playing a minimax game. Its extension, the conditional GANs \cite{Mirza2014Conditional}, fulfils the challenging task of generating images from abstract semantic labels. Li and Wand \cite{Li2016Precomputed} further showed that their Markovian GANs has certain advantages over the Gram-matrix-based methods \cite{gatys2016image,ulyanov2016texture} in coherent texture preservation.

\section{Proposed Method}

In this section, we first formulate our text effects transfer problem. Visual analytic is then presented on our observation of the high correlation between patch patterns (\emph{i.e.}~\textit{color} and \textit{scale}) and their spatial distributions in text effects images (Sec.~\ref{sec:analysis}). Based on this observation, we extract text effects statistics from the source images (Sec.~\ref{sec:statistics}) and employ it to adapt the texture synthesis algorithm for high-quality text effects transfer (Sec.~\ref{sec:transfer}).

\subsection{Problem Formulation and Analysis}
\label{sec:analysis}

Text effects transfer takes as input a set of three images, the source raw text image $S$, the source stylized image $S'$ and the target raw text image $T$, then automatically produces the target stylized image $T'$ with the text effects such that $S:S'::T:T'$. 

It is a quite challenging task to transfer arbitrary text effects automatically, due to the variety of text effects, the complex composition of sub-effects and the simpleness of guidance maps. To address this problem, we investigate the preferable text effects on the following two aspects: (i) how to determine the essential characteristics of text effects and (ii) how to characterize them mathematically. We start with a basic observation on text effects that the patch patterns are highly dominated by their locations. We develop to represent the pattern of a patch by two optimal factors: the pixel \textit{color} and the patch \textit{scale}. As shown intuitively in Figs.~\ref{fig:analysis}(a)-(d), the patches at similar locations (marked with the same color) tend to have similar patterns.

%To transfer arbitrary text effects automatically, we have to face a variety of text effects, the complex composition of text sub-effects and the simpleness of guidance maps. Thus, it is a quite challenging task. How to synthesize preferable text effects is still an open question. That is, (i) what is the essential factor to make text effects preferable? and (ii) how can we capture these characteristics for synthesis?

%Fortunately, after mathematically analyzing dozens of text effects created by designers, we find the clue: \textit{the high correlation between patch patterns (i.e.~color and scale) and their distances to text skeletons}. We note that textures with similar distances to the text skeleton tend to share similar patterns. It is schematically illustrated in Figs.~\ref{fig:analysis}(b)(d) where the patches with the same distance to the skeleton (in white) are marked by the same color. %It is mainly because designers usually adjust textures to fit the character shape for readability. This observation is consistent with the texture adaptation based on text shapes for readability conducted by designers in the real world.%the designers' adjustment on textures to fit the character shape for readability.

To quantitatively evaluate the locations of patches, we divide a text effects image into $N=16$ classes, namely, $N$ partitions. The modes of partition are extremely diverse and thus it is impractical to compare all of them. In this work, we compare five typical partition modes: (i) random: all pixels are randomly divided into $N$ equal partitions; (ii) grid: all partitions are evenly distributed according to their horizontal and vertical coordinates on the image; (iii) angle: all partitions are evenly distributed according to their angular coordinate, where the center of polar coordinate system is at the geometric center of the image; (iv) ring: all partitions are evenly distributed according to their radial coordinate, where the center of polar coordinate system is at the geometric center of the image; and (v) distance: all partitions are evenly distributed according to their geometric distance (the distance calculation will be given in Sec.~\ref{sec:distance}) to the skeleton of text on the image. In Fig.~\ref{fig:analysis}(e), the partitions modes of grid, angle, ring and distance have been intuitively illustrated, where all partitions are tinged differently.

%To quantitatively verify this correlation, we divide pixels/patches in text effects images into $N$ classes based on their distance to the text skeleton (distance calculation will be given in Sec.~\ref{sec:distance}) and use the differentiation of $N$ partitions as the measurement. Two examples of the distance-based partition are shown in the top row of Fig.~\ref{fig:analysis}(e), where we exploit different colors to denote each class.

%For the pixel color, taking \textit{flame} image for example, by marking each point in RGB space (Fig.~\ref{fig:analysis}(f)) with its class-color, we note that the points in Fig.~\ref{fig:analysis}(g) with the same class-color appear in the neighborhood, which means pixel colors have strong correlation with their distance values. Thus, we quantify it as the classification accuracy,
Then for each partition mode, we investigate the relationship between these partitions and the distributions of corresponding patterns. For the factor of \textit{color}, we represent its reliability by its classification accuracy of partitions:
\begin{equation}\label{eq:distance}
r_{\text{color}}=1-\epsilon,
\end{equation}
where $\epsilon$ is the training error or empirical risk obtained by training SVM~\cite{CC01a} to classify the color given a type of partition. We have tested on $30$ text effects images created by designers to obtain their reliability on color classification. The average reliability are then shown in Table~\ref{tb:table}, where only the relative values are instructive in our design. From this table, the distance is demonstrated to be the most reliable factor to depict pixel colors, with a value of $0.147$ on average. In Fig.~\ref{fig:analysis}(g), pixels of the \textit{flame} image are tinged according to their distance (see the top left image of Fig.~\ref{fig:analysis}(e)) in RGB space. We note that the points with the same class-color appear in the neighborhood. It is also intuitively shown that the color and distance are highly correlated in text effects.

%where $\epsilon$ is the training error or empirical risk obtained by training SVM~\cite{CC01a} to classify pixel colors given our distance-based partition. The mean correlation between pixel colors and the distance values on $30$ text effects images created by designers is $0.147$. We also provide other three partition modes (as shown in the bottom row of Fig.~\ref{fig:analysis}(e)) as well as the random partition mode for comparison in the experiments. Their mean correlations with pixel colors over $30$ test images are shown in the second row of Table~\ref{tb:table}. As expected, the distance is the most important factor to depict pixel colors.

The distance has also shown its importance in characterizing the scale of patterns. Firstly, for different patch sizes, we calculate the average patch difference between all patches in a partition and their best matches on the same image, which forms a response curve of scale. Then, for all the $N$ partitions with the same partition mode, we have $N$ response curves that show the impacts of scales. Two examples of response curves for \textit{denim fabric} image are shown in Figs.~\ref{fig:analysis}(h) and (i), where each point shows its average and standard deviation of patch differences under the same partition and scale. To compare the reliability of all partition modes, two terms are utilized: (i) inter curve standard deviation $\sigma_{\text{inter}}$: the average of the scale-wise standard deviations of average responses at same partitions; and (ii) intra curve standard deviation $\sigma_{\text{intra}}$: the average of point-wise standard deviations for all scales and partitions. A higher $\sigma_{\text{inter}}$ implies that sub-effects are easier to be distinguished by their locations, while a lower  $\sigma_{\text{intra}}$ implies patches in the same partition react uniformly to scale changing and possibly share common optimal scale for description. Therefore, we evaluate the reliability by
%For the patch scale, we enumerate patch sizes and calculate the difference between each patch and its best match, which forms a response curve at different scaling. Taking \textit{denim fabric} image for example, Figs.~\ref{fig:analysis}(h) and (i) show $N$ response curves of  patch sizes in distance and random modes, respectively. Each point on the curve gives the mean and standard deviation of patch differences at a certain patch size in the corresponding class. We find that: (i) Response curves in distance mode are more diverse (high inter-curve standard deviations $\sigma_{\text{inter}}$), which means different sub-effects are well distinguished by their distance values. (ii) Points on response curves in distance mode have lower standard deviations (low intra-curve standard deviations $\sigma_{\text{intra}}$), which means patches with similar distances react uniformly to scale changing and possibly share common optimal scale for description. Considering these two aspects, we evaluate the correlation with the patch scale by,
\begin{equation}\label{eq:scales}
r_{\text{scale}}=\sigma_{\text{inter}}/\sigma_{\text{intra}}.
\end{equation}
%Here $\sigma_{\text{inter}}$ and $\sigma_{\text{intra}}$ are defined as sum of standard deviations $\sigma$ with different patch sizes. Given a patch size, $\sigma$ is standard deviation of mean patch differences of $N$ points in different curves for $\sigma_{\text{inter}}$ and is mean of standard deviations of $N$ points for $\sigma_{\text{intra}}$. The bottom row of Table~\ref{tb:table} shows the mean correlations between the patch scale and different modes on $30$ images where distance owns the highest value.
The reliability of all the five partition modes are then given in Table~\ref{tb:table} where the factor of distance achieves highest to characterize the patch scales.

%Based on the results for pixel colors and patch scales, we obtain the conclusion that the high correlation between patch patterns and their distances is reasonable essential characteristics for high-quality text effects.
As a conclusion, there exist \textit{high correlations between patch patterns (i.e.~color and scale) and their distances to text skeletons}. These are reasonable essential characteristics for high-quality text effects.

%For \textit{denim fabric} example, correlations between patch size and spatial distribution \textit{w.r.t.} distance/ring/grid/angle/random are $3.104/1.683/0.592/0.284/0.108$, respectively. This verifies the distance is the most important factor to determine patch scales.

\begin{table} [t]
\caption{Reliability between patch patterns and different partitions.}\vspace{1mm}
\label{tb:table}
\centering
\begin{tabular}{c|c|c|c|c|c}
\hline
$r$ & rand & gird & angle & ring & dist\\
\hline
color & 0.063 & 0.106 & 0.119 & 0.105 & \textbf{0.147}\\
\hline
scale & 0.153 & 0.793 & 0.486 & 0.590 & \textbf{0.950}\\
\hline
\end{tabular}
\vspace{-4mm}
\end{table}

%Now that we have answered our first question on the essential characteristics for text effects. In the following, we will give our solution to our second question on the modelling and utilization of these characteristics.

\subsection{Text Effects Statistics Estimation}
\label{sec:statistics}

We now convert the aforementioned analysis into patch statistics that can be directly used as the transfer guidance. %Specifically, we detect the optimal scales for source patches, and estimate their normalized distances to the text skeleton. Then we are able to derive the posterior probability of the optimal scale for each patch based on its spatial position.
For our patch-based algorithm, in the following we use $p$ and $q$ to denote the pixels in $T/T'$ and $S/S'$, respectively, and use $P(p)$ and $P'(p)$ to represent the patches centered at $p$ in $T$ and $T'$, respectively. The same goes for patches $Q(q)$ and $Q'(q)$ in $S$ and $S'$.

\begin{figure}[t]
  \centering
    \subfigure[Optimal scale map]{
    \includegraphics[width=0.45\linewidth]{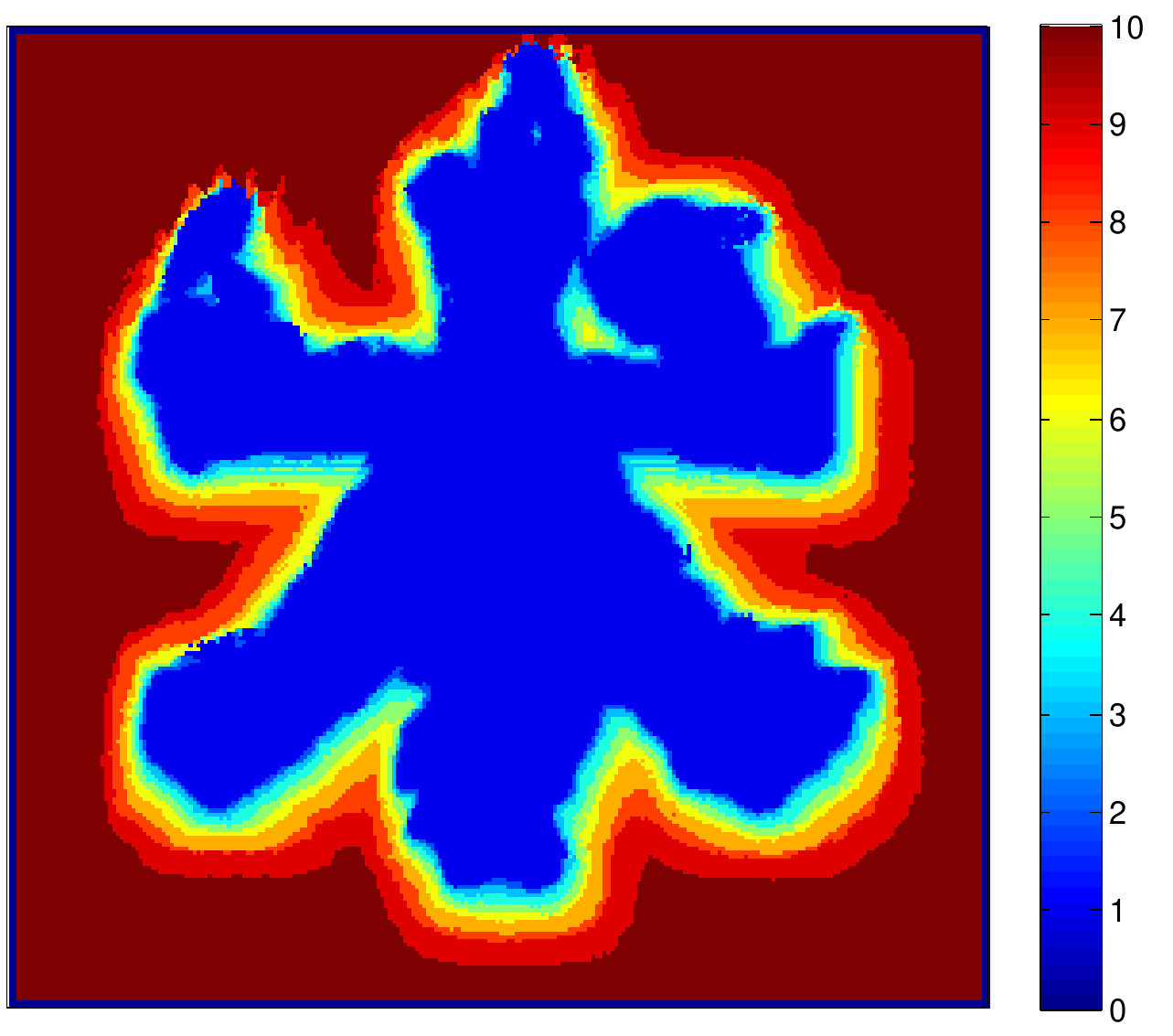}}
    \subfigure[Visualized patch scale]{
    \includegraphics[width=0.39\linewidth]{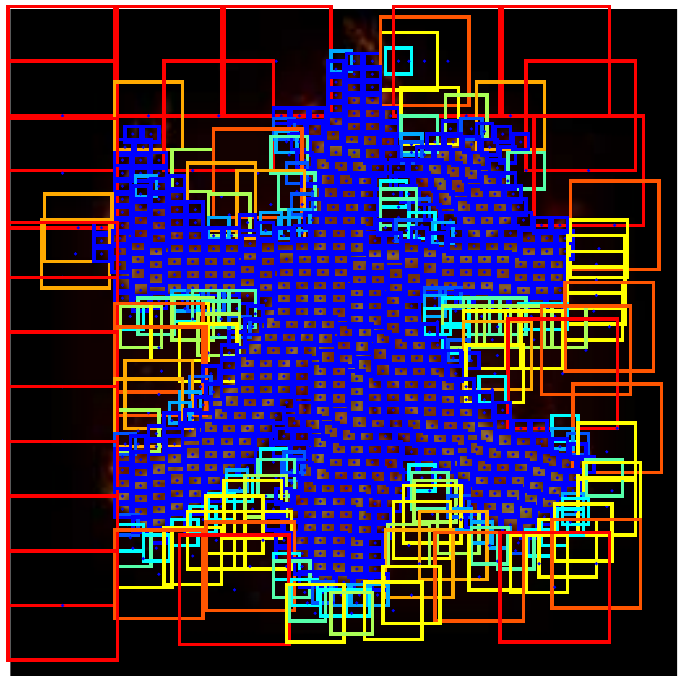}}
  \caption{Detected optimal patch scales for the \textit{flame} image.}
  \label{fig:scale}
  \vspace{-3mm}
\end{figure}
\vspace{-2mm}
\subsubsection{Optimal Patch Scale Detection}
\vspace{-1mm}
Inspired by \cite{Frigo2016Split}, we propose a simple yet effective approach to detect the optimal patch scale $\text{scal}(q)$ to depict texture patterns round $q$. Given a predefined downsample factor $s$, we start from the max (roughest) scale $L$ to filter source patches and let the screened patches pass to a finer scale.

We use a fixed patch size of $m\times m$ and resize the image to accomplish multiple scales. Let $S_\ell$ be the downsampled source $S$ with a scale rate of $1/s^{\ell-1}$ and $Q_\ell(q)$ be the patch centered at $q/s^{\ell-1}$ in $S_\ell$. $S'_\ell$ and $Q'_\ell(q)$ are similarly defined. If $\hat{q}$ is the correspondence of $q$ at scale $\ell$ such that
\begin{equation}\label{eq:distance}
\hat{q}=\arg\min\|Q_\ell(q)-Q_\ell(\hat{q})\|^2+\|Q'_\ell(q)-Q'_\ell(\hat{q})\|^2,
\end{equation}
then our filter criterion at scale $\ell$ is
\begin{equation}\label{eq:split}
\zeta_\ell(q,\hat{q})=\big(\sigma_{\ell}+\sqrt{d_\ell(q,\hat{q})}>\omega\big),
\end{equation}
where $\sigma_{\ell}=\sqrt{\text{\textit{Var}}(Q'_\ell(q))}/2$. Patches that satisfy the filter criterion  pass through to finer scale $\ell-1$, while the filter residues set  $\ell$ as their optimal scales.
The optimal patch scale detection is summarized in Algorithm \ref{algorithm}.
An example of the optimal scales for the \textit{flame} image is shown in Fig.~\ref{fig:scale}(a). It is found that the textured region near the character requires finer patch scales than the outer flat region. For better visualization, we show the optimal scale of the patch $Q(q)$ by resizing it at a scale rate of $s^{\text{scal}(q)-1}$ in Fig.~\ref{fig:scale}(b).
%\iffalse
\begin{algorithm}
  \caption{Optimal Patch Scale Detection}
  \textbf{Input:} Image $S,S'$, parameters $L,s,\omega$\\
  \textbf{Output:} Optimal scale $\text{scal}(q)$ for each pixel $q$
  \begin{algorithmic}[1]
    \State Initialize $R=\{q|q\in S\}$ and $\text{scal}(q)=1,\forall q\in R$
    \For {$\ell=L,...,2$}
        \ForAll {$p\in R$}
            \State Compute $\hat{q}=\arg\min_{\hat{q}} d_\ell(q,\hat{q})$
            \If {$\zeta_\ell(q,\hat{q})$ is false}
                \State $\text{scal}(q)=\ell$
                \State $R=R\setminus\{q\}$
            \EndIf
        \EndFor
    \EndFor
  \end{algorithmic}
  \label{algorithm}
\end{algorithm}
%\fi
\subsubsection{Robust Normalized Distance Estimation}
\label{sec:distance}
\begin{figure}
  \centering
    \subfigure[Text image]{
    \includegraphics[width=0.3\linewidth]{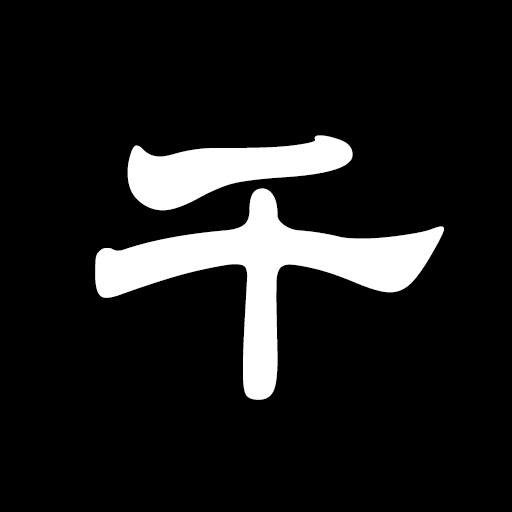}}
    \subfigure[Notation definition]{
    \includegraphics[width=0.61\linewidth]{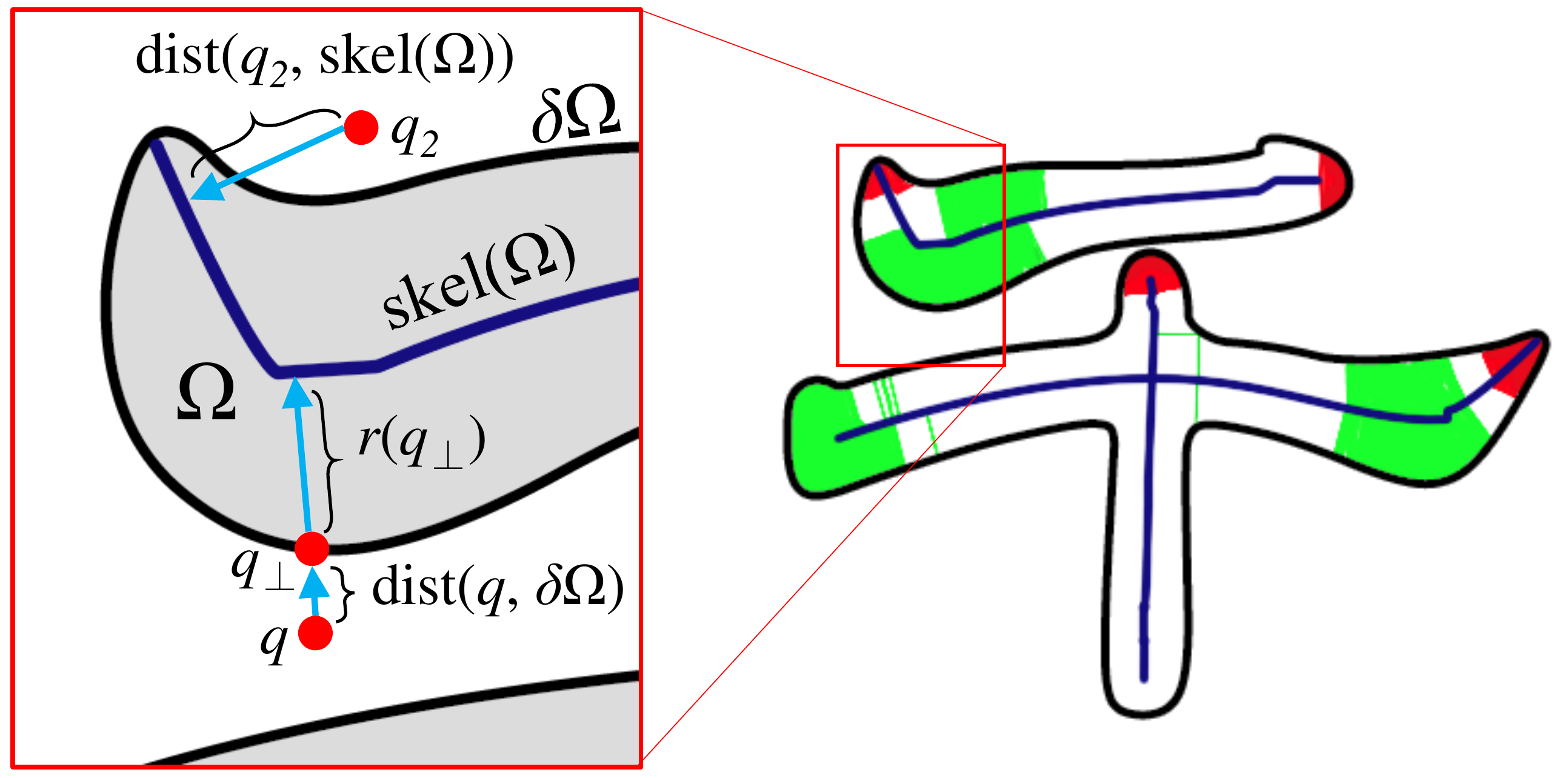}}\vspace{-3mm}
    \subfigure[$\text{d}\widetilde{\text{is}}\text{t}(q,\text{skel}(\Omega))$]{
    \includegraphics[width=0.36\linewidth]{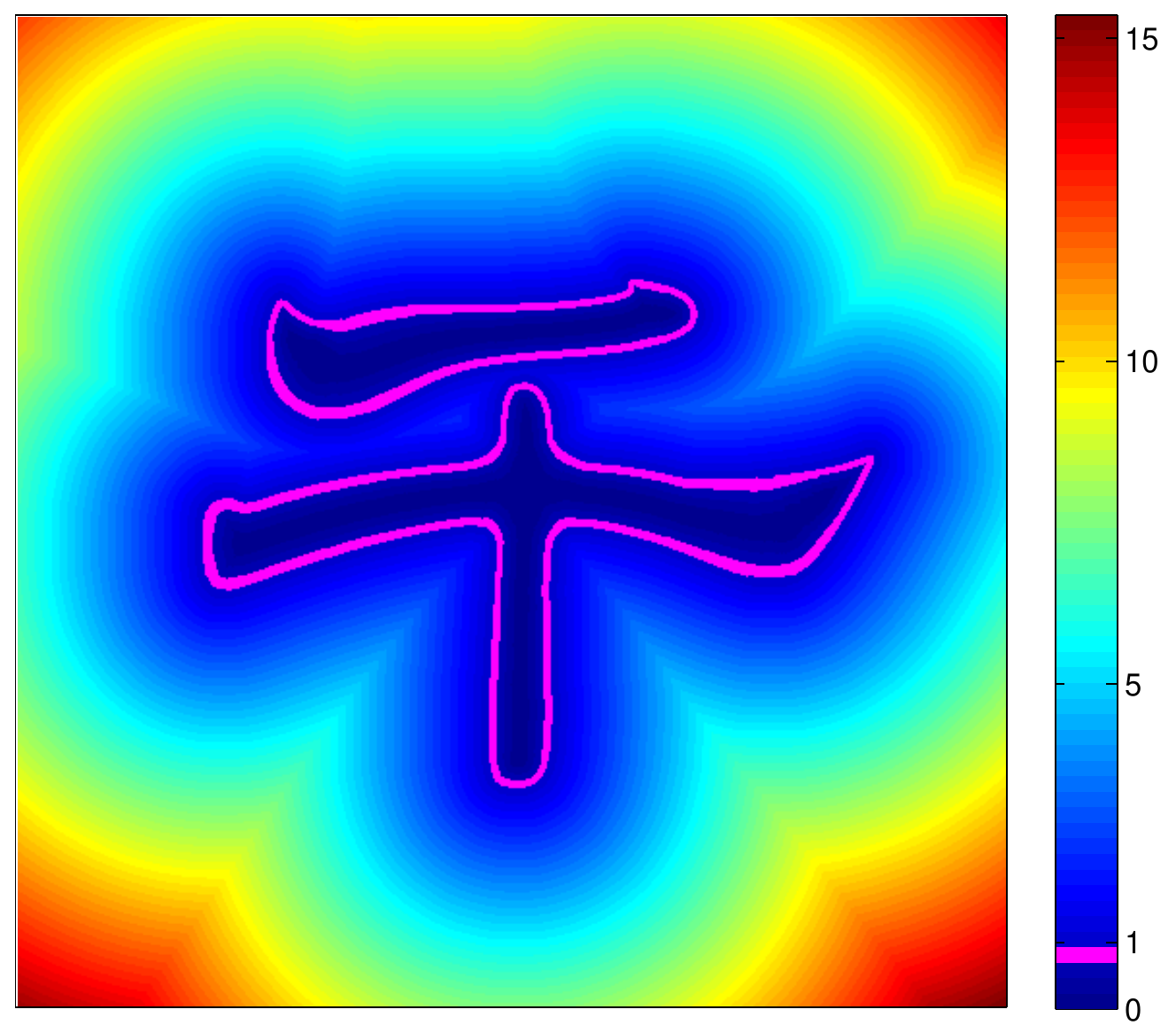}}
    \subfigure[Linear relation of $r(q)$ and $rank(q)$]{
    \includegraphics[width=0.56\linewidth]{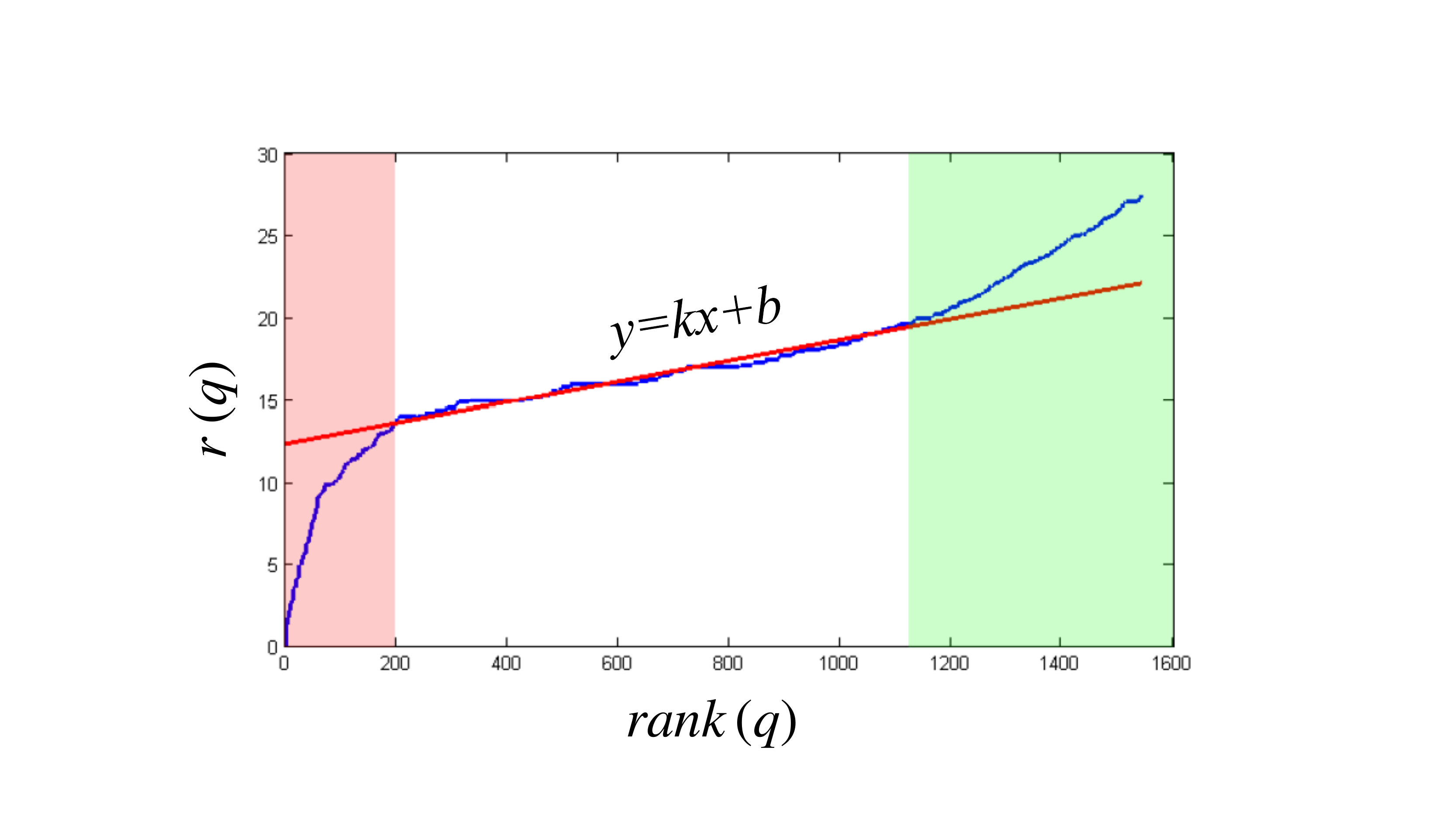}}
  \caption{Robust normalized distance estimation. (a) The text image. (b) Our detected text skeleton and the notation definition. (c) The estimated normalized distance. The distance of the pixels on the text boundary to the text skeleton are normalized to $1$ (colored by magenta). (d) The statistics of the text width. }\label{fig:dist}
  \vspace{-2mm}
\end{figure}

\vspace{-0.5mm}
Here we first define some concepts. In the text image, its text region is denoted by $\Omega$. The skeleton $\text{skel}(\Omega)$ is a kernel path within $\Omega$. We use $\text{dist}(q,A)$ to denote the distance between $q$ and its nearest pixel in set $A$. We are going to calculate $\text{dist}(q,\text{skel}(\Omega))$. For $q$ on the text contour $\delta\Omega$, the distance is also known as the text width or radius $r(q)$. Fig.~\ref{fig:dist}(b) gives the visual interpretation.

We extract $\text{skel}(\Omega)$ from $S$ using morphology operations. To ensure the distance invariant to the text width, we aim to normalize the distance so that the normalized text width equals to $1$. Simply dividing the distance by the text width is unreliable because the inaccurate of the obtained $\text{skel}(\Omega)$
%makes both the numerator and denominator inaccuracy as well.
leads to errors both in the numerator and denominator as well.
To address this issue, we estimate corrected text width $\widetilde{r}(q)$ based on statistics and use the accurate $\text{dist}(q,\delta\Omega)$ to derive normalized $\text{d}\widetilde{\text{is}}\text{t}(q,\text{skel}(\Omega))$.

Specifically, we sort $r(q), \forall q\in\delta\Omega$ and obtain their rankings $rank(q)$. We observe that the relation between $r(q)$ and $rank(q)$ can be well modelled by linear regression, as shown in Figs.~\ref{fig:dist}(d). From Figs.~\ref{fig:dist}(b)(d), we discover that outliers assemble at small values. We empirically assume the leftmost $20\%$ points are outliers and eliminate them by
\begin{equation}\label{eq:distance}
\widetilde{r}(q)=\max(\text{dist}(q,\text{skel}(\Omega)),0.2k|\delta\Omega|+b),
\end{equation}
where $k,b$ are linear regression coefficients, $|\delta\Omega|$ is the pixel number of $\delta\Omega$. Finally, the normalized distance is obtained,
\begin{equation}\label{eq:distance}
\text{d}\widetilde{\text{is}}\text{t}(q,\text{skel}(\Omega))=\left\{
\begin{aligned}
& 1+\text{dist}(q,\delta\Omega)/\overline{r},\text{ if } q\notin\Omega\\
& 1-\text{dist}(q,\delta\Omega)/\widetilde{r}(q_{\bot}),\text{other}\\
\end{aligned}
\right.,
\end{equation}
where $q_{\bot}\in\delta\Omega$ is the nearest pixel to $q$ along $\delta\Omega$ and $\overline{r}=0.5k|\delta\Omega|+b$ is the mean text width.

For simplicity, we omit $\text{skel}(\Omega)$ and use $\text{dist}(q)$ to refer to $\text{d}\widetilde{\text{is}}\text{t}(q,\text{skel}(\Omega))$ in the following.

\subsubsection{Optimal Scale Posterior Probability Estimation}
\label{sec:dist-scale}

%Our key insight is that the good text effects share strong correlations between their patterns and spatial distributions.
%Now that we have estimated distances as well as the optimal scales of the patches throughout the source image, we can model the correlation using statistics.
In this section, we derive the posterior probability of the optimal patch scale to model the aforementioned high correlation between patch patterns and their spatial distributions.

We uniformly quantify all distances into $100$ bins and denote $\text{bin}(q)$ as the bin $q$ belongs to. Then, a 2-d histogram $hist(\ell,x)$ is computed:
\begin{equation}
hist(\ell,x)=\sum_{q}\psi(\text{scal}(q)=\ell\wedge\text{bin}(q)=x),
\end{equation}
where $\psi(\cdot)$ is $1$ when the argument is true and $0$ otherwise. And the joint probability of the distance and the optimal scale can be estimated as,
\begin{equation}
\mathcal{P}(\ell,x)=hist(\ell,x)/\sum_{\ell,x}hist(\ell,x).
\end{equation}
Finally, the posterior probability $\mathcal{P}(\ell|\text{bin}(q))$ for $\ell$ being the appropriate scale to depict the patches with distances corresponding to $\text{bin}(q)$ can be deduced:
\begin{equation}\label{eq:weight}
\mathcal{P}(\ell|\text{bin}(p))=\mathcal{P}(\ell,\text{bin}(p))/\sum_\ell \mathcal{P}(\ell,\text{bin}(p)).
\end{equation}

We assume the target images share the same posterior probability with the source image. And we will use this probability to select patch scales statistically for texture synthesis to adapt extremely various text effects.

\subsection{Text Effect Transfer}
\label{sec:transfer}

In this section, we describe how we adapt conventional texture synthesis method to dealing with the challenging text effects. We build on the texture synthesis method of Wexler \textit{et al.}~\cite{Space-time} and its variants~\cite{ImageMelding} using random search and propagation as in PatchMatch~\cite{PatchMatch,GeneralizedPatchMatch}. We refer to these papers for details of the base algorithm.

% 我们修改方程加入对文字结构的约束，构成了我们的baseline。 然后我们在basline的基础上增加weight对结构和纹理施加多尺度。随后增加使分布一致。最后，加入。。。使结果符合人的心里视觉。

We apply character shape constrains to the patch appearance measurement to build our baseline, and further incorporate estimated text effects statistics to accomplish adaptive multi-scale style transfer (Sec.~\ref{sec:app-term}). Then a distribution term is introduced to adjust the spatial distribution of the text sub-effects (Sec.~\ref{sec:dist-term}). Finally, we propose a psycho-visual term that prevents texture over-repetitiveness for naturalness (Sec.~\ref{sec:psy-term}).

\subsubsection{Objective Function}

We augment the texture synthesis objective function in \cite{Space-time} by including a distribution term and a psycho-visual term. And our objective function takes the following form,
\begin{equation}\label{eq:objective_function}
\min_{q}\sum_{p}E_{\text{app}}(p,q)+\lambda_1E_{\text{dist}}(p,q)+\lambda_2E_{\text{psy}}(p,q),\vspace{-1mm}
\end{equation}
where $p$ is the center position of a target patch in $T$ and $T'$, $q$ is the center position of the corresponding source
patch in $S$ and $S'$. The three terms $E_{\text{app}}$, $E_{\text{dist}}$ and $E_{\text{psy}}$ are the appearance, distribution and psycho-visual terms, respectively, which are weighted by $\lambda_1$ and $\lambda_2$ to together make up the patch distance.

\subsubsection{Appearance Term: Texture Style Transfer}
\label{sec:app-term}

The original texture synthesis algorithm of Wexler \textit{et al.} \cite{Space-time} minimizes the Sum of the Squared Differences (SSD) of two patches sampled from texture image pair $S'/T'$. We adapt it to texture transfer tasks by applying additional SSD of two patches sampled from the text image pair $S/T$:
\begin{equation}\label{eq:baseline-function}
E_{\text{app}}(p,q)=\lambda_{3}\|P(p)-Q(q)\|^2+\|P'(p)-Q'(q)\|^2,
\end{equation}
where $\lambda_{3}$ is a weight that compromises between the color difference and character shape difference. %Note that during the optimization process, $P$ updates after each iteration and $P'$ remains constant.
We take the objective function that only minimizes the appearance term in Eq. (\ref{eq:baseline-function}) as our baseline.

Stylized texts often contain multiple sub-effects with different optimal representation scales. Thus, in addition to the baseline, we propose the adaptive scale-aware patch distance by incorporating the estimated posterior probability,

\begin{equation}\label{eq:app_function2}
\begin{aligned}
E_{\text{app}}(p,q)=&\lambda_{3}\sum_{\ell}\mathcal{P}(\ell|\text{bin}(p))\|P_\ell(p)-Q_\ell(q)\|^2\\
+&\sum_{\ell}\mathcal{P}(\ell|\text{bin}(p))\|P'_\ell(p)-Q'_\ell(q)\|^2.
\end{aligned}
\vspace{-1mm}
\end{equation}
The posterior probability helps to explore patches through multiple appropriate scales for better textures synthesis.

\subsubsection{Distribution Term: Spatial Style Transfer}
\label{sec:dist-term}

The distribution of sub-effects highly correlates with their distances to the text skeleton. Based on this prior, we introduce a distribution term,
\begin{equation}\label{eq:dist_function}
E_{\text{dist}}(p,q)=\big(\text{dist}(p)-\text{dist}(q)\big)^2/\max(1,\text{dist}^2(p)),
\end{equation}
which encourages the text effects of the target to share similar distribution with the source image, thereby realizing a spatial style transfer. To ensure that the cost is invariant to the image scale, we add the denominator $\max(1,\text{dist}^2(p))$.

\subsubsection{Psycho-Visual Term: Naturalness Preservation}
\label{sec:psy-term}

Texture over-repetitiveness can seriously reduce human subjective evaluation in the aesthetics. Therefore,
%while for each target patch we try to find its best matched source patch,
we aim to penalize certain source patches to be selected repetitiously.

Let $\Phi(q)$ be the set of pixels that currently finds $q$ as its correspondence and $|\Phi(q)|$ be the size of the set. We define the psycho-visual term as,
\begin{equation}\label{rep_function}
E_{\text{psy}}(p,q)=|\Phi(q)|.
\end{equation}
From the perspective of $q$, we can better understand this repetitiveness penalty:
\begin{equation}\label{eq:rep_function2}
\begin{aligned}
%\sum_{p}E_{\text{psy}}(p,q)&=\sum_{p}|\Phi(q)|=\sum_{q}\sum_{p\in\Phi(q)}|\Phi(q)|\\&=\sum_{q}|\Phi(q)|^2.
\sum_{p}|\Phi(q)|=\sum_{q}\sum_{p\in\Phi(q)}|\Phi(q)|=\sum_{q}|\Phi(q)|^2.
\end{aligned}
\end{equation}
Since $\sum_{q}|\Phi(q)|=|T|$ is constant, Eq.~(\ref{eq:rep_function2}) reaches the minimum when all $|\Phi(q)|$ equals. It means our psycho-visual term encourages source patches to be used evenly.

\subsubsection{Function Optimization}

We follow the iterative coarse-to-fine matching and voting steps as in \cite{Space-time}. In the matching step, PatchMatch algorithm \cite{PatchMatch,GeneralizedPatchMatch} is adopted. We update $\Phi(q)$ after each iteration of search and propagation for the psycho-visual term. Meanwhile, the initialization of $T'$ plays an important role in the final results, since our guidance map provides very few constraints on textures. We vote the source patches that are searched to only minimize Eq.~(\ref{eq:dist_function}) to form our initial guess of $T'$. This simple strategy improves the final results significantly as shown in Fig.~\ref{fig:ana-dist}.
% 初始化

\section{Analysis}
\label{sec:overall-analysis}
\vspace{-1mm}
\begin{figure}
  \centering
    \subfigure[$m=5,L=1$]{
    \includegraphics[width=0.3\linewidth]{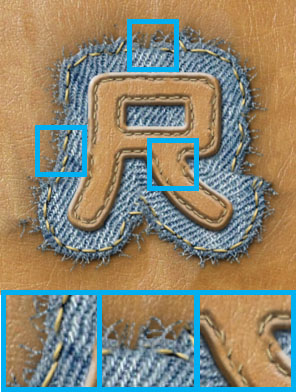}}
    \subfigure[$m=15,L=1$]{
    \includegraphics[width=0.3\linewidth]{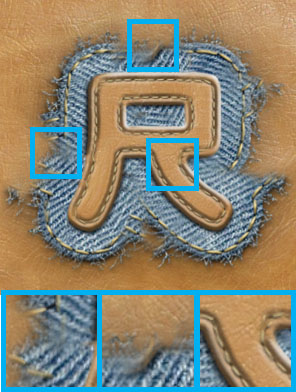}}
    \subfigure[$m=5,L=5$]{
    \includegraphics[width=0.3\linewidth]{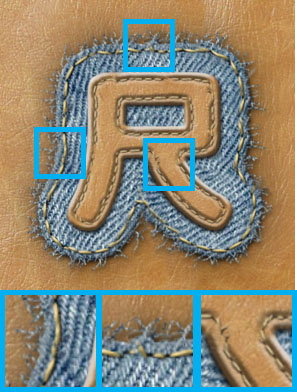}}
  \caption{Effects of the multi-scale strategy. (a) Results using single-scale $5\times5$ patches. (b) Results using single-scale $15\times15$ patches. (c) Results using joint $5\times5$ patches over $5$ scales.}\label{fig:ana-scale}
  \vspace{-4mm}
\end{figure}

%This section analyzes the advantages of the proposed three terms.

\begin{figure}
  \centering
    \subfigure[$\lambda_1=0.0$]{
    \includegraphics[width=0.3\linewidth]{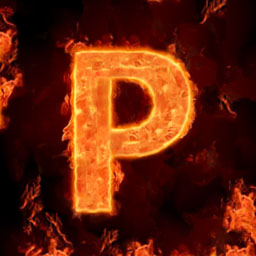}}
    \subfigure[$\lambda_1=0.1$]{
    \includegraphics[width=0.3\linewidth]{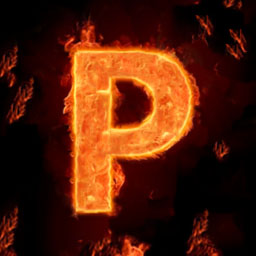}}
    \subfigure[$\lambda_1=0.1 +$ Init]{
    \includegraphics[width=0.3\linewidth]{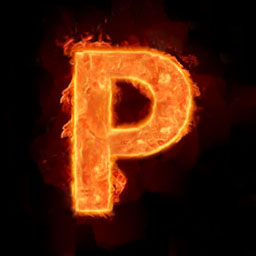}}
  \caption{Effects of the distribution term. (a) Results without distribution term. (b) Results obtained by random initialization and optimization with distribution term. (c) Results obtained by both initialization and optimization with distribution term.}\label{fig:ana-dist}
  \vspace{-5mm}
\end{figure}

\begin{figure}
  \centering
    \subfigure[$\lambda_2=0.0$]{
    \includegraphics[width=0.3\linewidth]{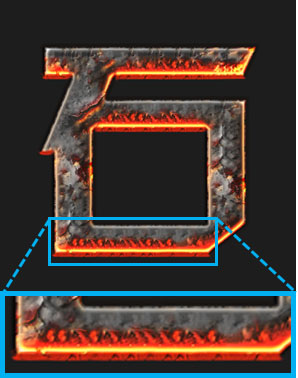}}
    \subfigure[$\lambda_2=0.005$]{
    \includegraphics[width=0.3\linewidth]{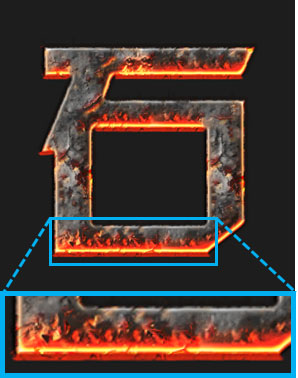}}
    \subfigure[$\lambda_2=0.01$]{
    \includegraphics[width=0.3\linewidth]{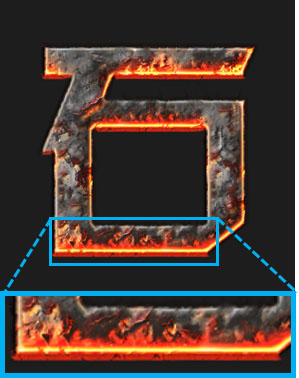}}
  \caption{Effects of the psycho-visual term, which penalizes texture over-repetitiveness and encourages new texture creation. }\label{fig:rep}
  \vspace{-4.5mm}
\end{figure}

\begin{figure*}[t]
\centering
\subfigure{
\includegraphics[width=0.132\linewidth]{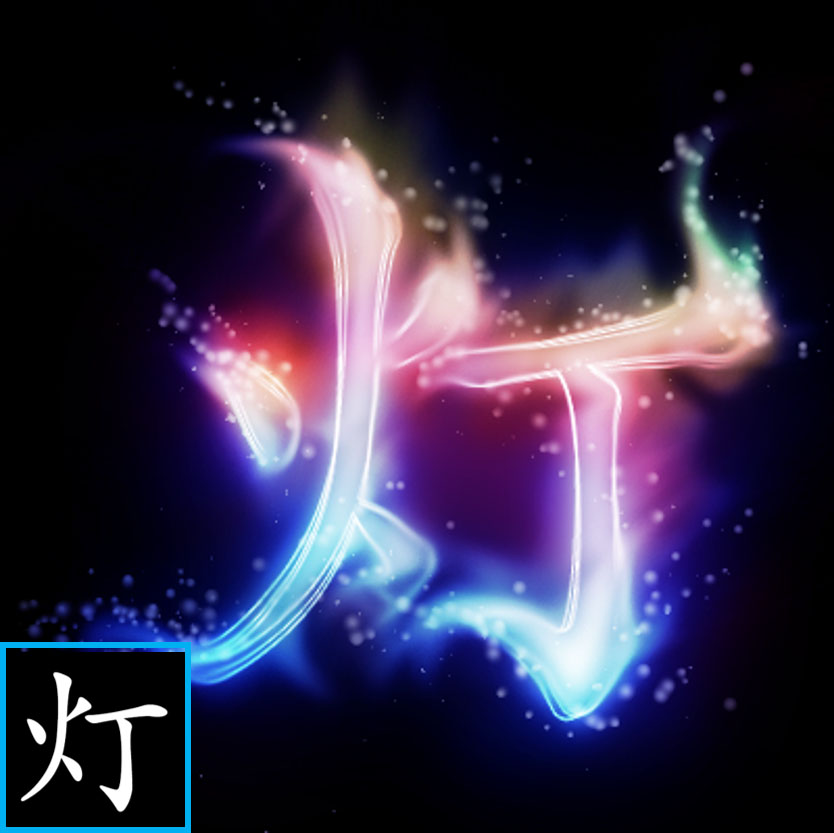}}
\subfigure{
\includegraphics[width=0.132\linewidth]{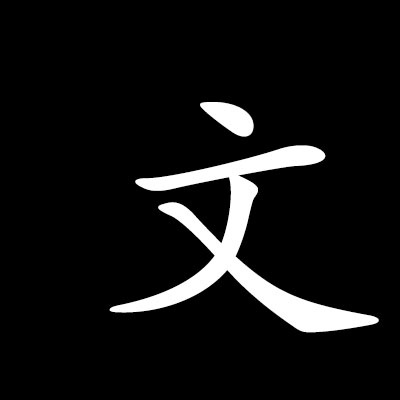}}
\subfigure{
\includegraphics[width=0.132\linewidth]{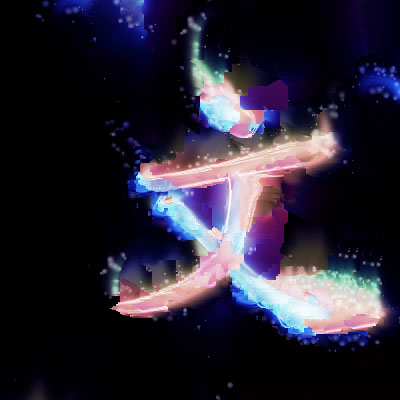}}
\subfigure{
\includegraphics[width=0.132\linewidth]{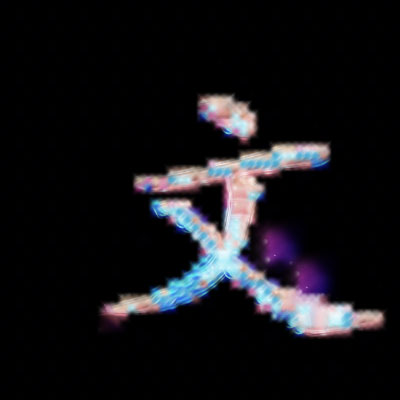}}
\subfigure{
\includegraphics[width=0.132\linewidth]{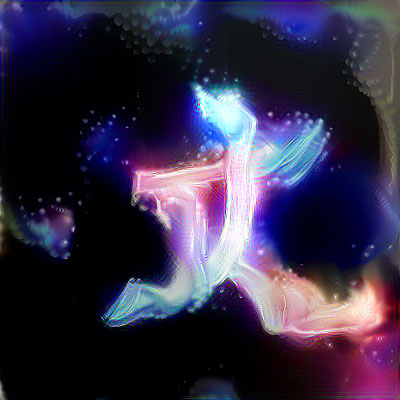}}
\subfigure{
\includegraphics[width=0.132\linewidth]{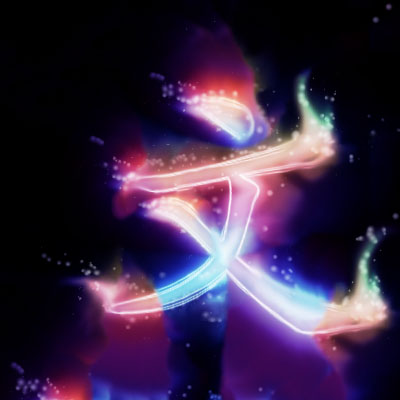}}
\subfigure{
\includegraphics[width=0.132\linewidth]{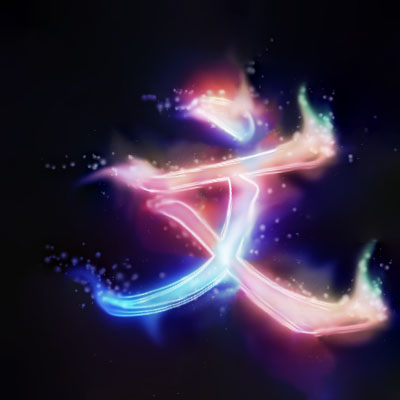}}\vspace{-2mm}
%\subfigure{
%\includegraphics[width=0.132\linewidth]{figures/yin-silver3.png}}
%\subfigure{
%\includegraphics[width=0.132\linewidth]{figures/jin-text.png}}
%\subfigure{
%\includegraphics[width=0.132\linewidth]{figures/jin-silver-yin-analogy.png}}
%\subfigure{
%\includegraphics[width=0.132\linewidth]{figures/invalid.png}}
%\subfigure{
%\includegraphics[width=0.132\linewidth]{figures/jin-silver-yin-doodles.png}}
%\subfigure{
%\includegraphics[width=0.132\linewidth]{figures/jin-silver-yin-bl.png}}
%\subfigure{
%\includegraphics[width=0.132\linewidth]{figures/jin-silver-yin-ours.png}}\vspace{-2mm}
\subfigure{
\includegraphics[width=0.132\linewidth]{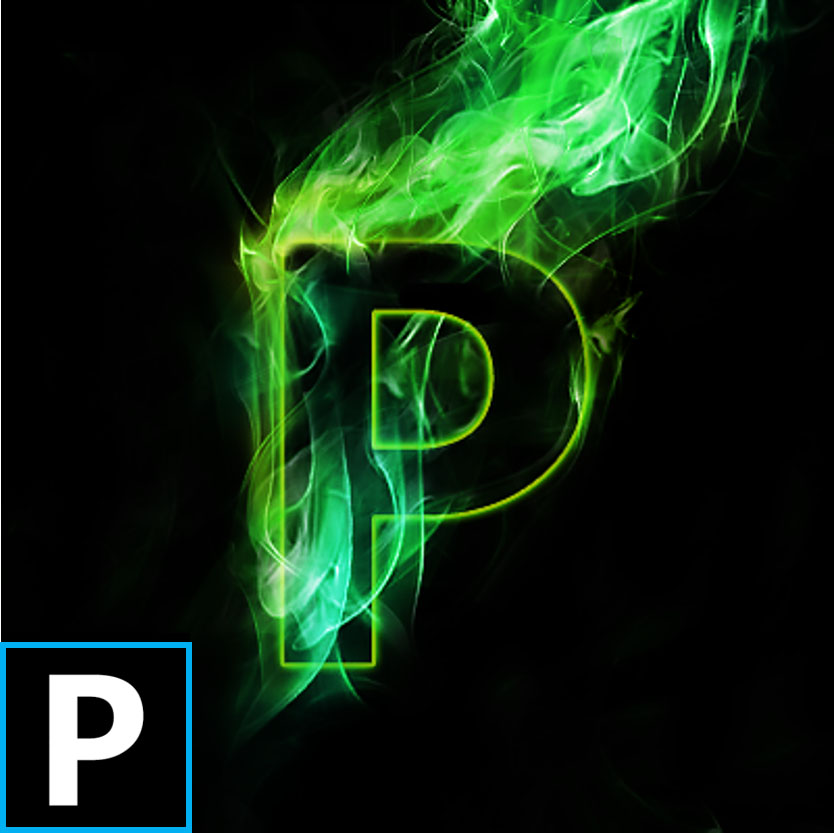}}
\subfigure{
\includegraphics[width=0.132\linewidth]{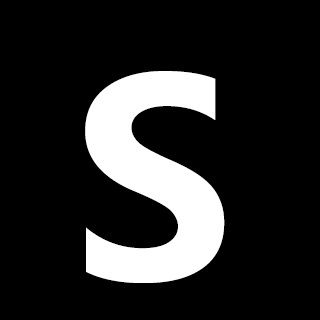}}
\subfigure{
\includegraphics[width=0.132\linewidth]{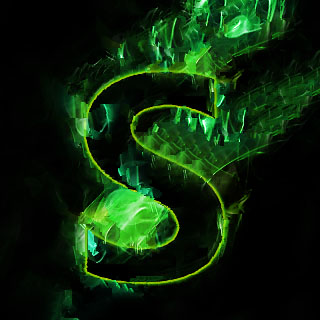}}
\subfigure{
\includegraphics[width=0.132\linewidth]{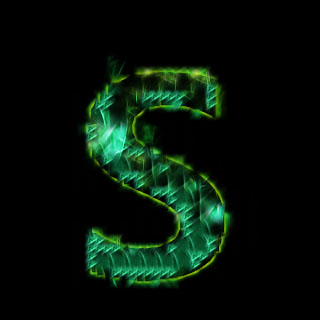}}
\subfigure{
\includegraphics[width=0.132\linewidth]{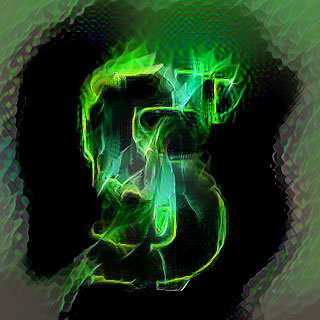}}
\subfigure{
\includegraphics[width=0.132\linewidth]{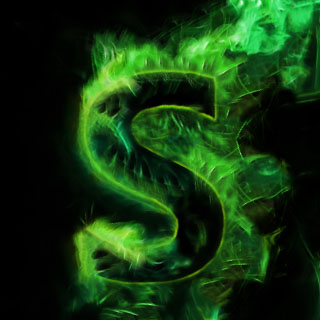}}
\subfigure{
\includegraphics[width=0.132\linewidth]{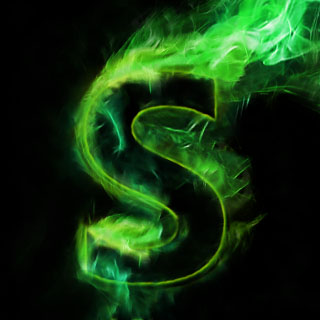}}\addtocounter{subfigure}{-14}\vspace{-2mm}
\subfigure[$(S,S')$]{
\includegraphics[width=0.132\linewidth]{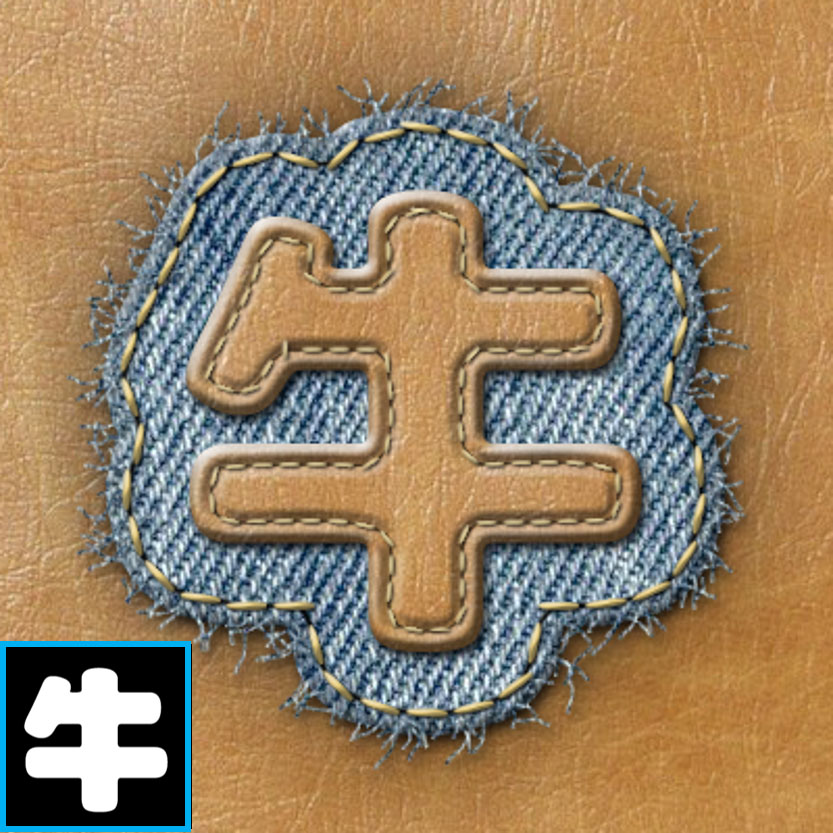}}
\subfigure[$T$]{
\includegraphics[width=0.132\linewidth]{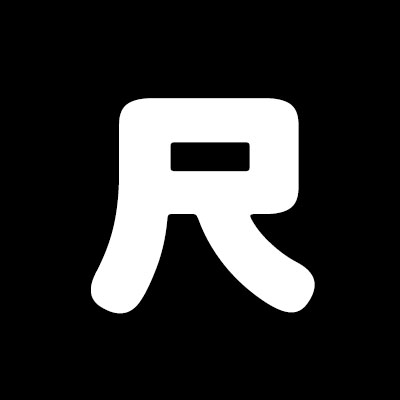}}
\subfigure[Analogy~\protect\cite{Hertzmann2001Image}]{
\includegraphics[width=0.132\linewidth]{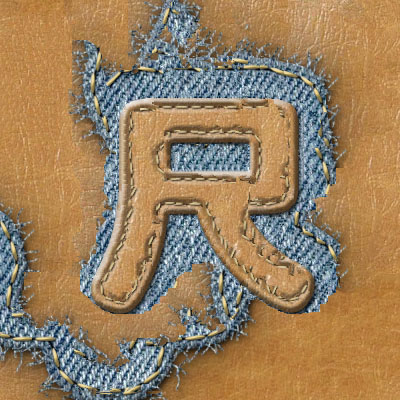}}
\subfigure[Split \& match~\protect\cite{Frigo2016Split}]{
\includegraphics[width=0.132\linewidth]{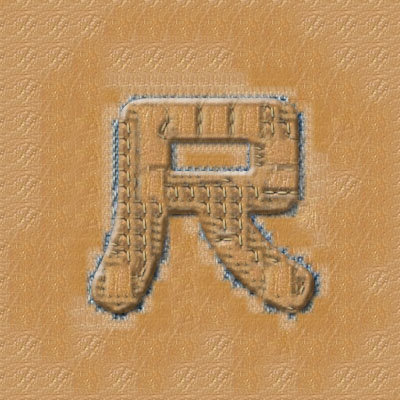}}
\subfigure[Neural doodle~\protect\cite{Champandard2016Semantic}]{
\includegraphics[width=0.132\linewidth]{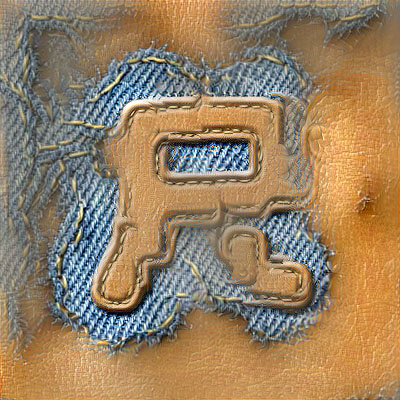}}
\subfigure[Baseline]{
\includegraphics[width=0.132\linewidth]{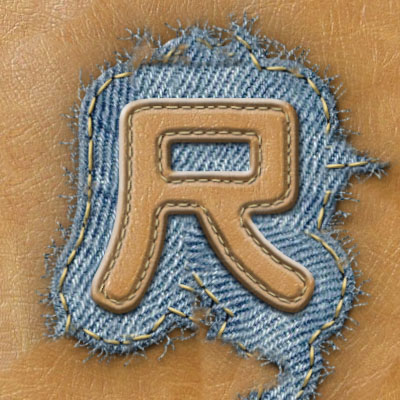}}
\subfigure[Proposed method]{
\includegraphics[width=0.132\linewidth]{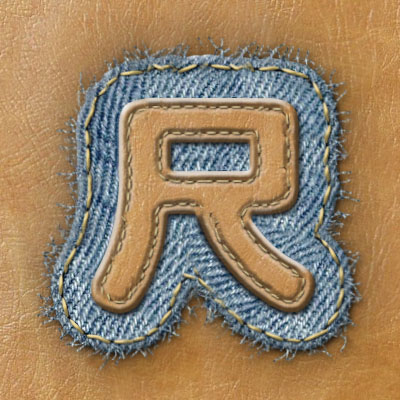}}
\caption{Comparison with state-of-the-art methods on various text effects. From top to bottom: \textit{neon},  \textit{smoke}, \textit{denim fabric}. (a) Input source text effects with their raw text counterparts in the lower-left corner. (b) Target text. (c) Results of Image Analogies \cite{Hertzmann2001Image}. (d) Results of Split and Match \cite{Frigo2016Split}.  (e) Results of Neural Doodles \cite{Champandard2016Semantic}. (f) Results of our baseline method. (g) Results of the proposed method.}
\label{fig:comparison}
\vspace{-4mm}
\end{figure*}

\textbf{Appearance Term.} The advantages of the proposed appearance term lie in two aspects: (i) Preserve coarse grained texture structures. (ii) Preserve texture details. We show in Figs.~\ref{fig:ana-scale}(a) and (b) the \textit{denim fabric} style generated using single-scale $5\times5$ and $15\times15$ patches, respectively. Small patches capture very limited contextual information, thus it cannot guarantee the structure continuity. As can be seen in Fig.~\ref{fig:ana-scale}(a), sewing threads look cracked and are not along the uniform directions. However, choosing large patches leads to smoothing out tiny thread residues as in Fig.~\ref{fig:ana-scale}(b). These problems are well solved by jointly using $5\times5$ patches over $5$ scales as in Fig.~\ref{fig:ana-scale}(c), where the overall shape is well preserved and the details like sewing threads look more vivid.

\textbf{Distribution Term.} The distribution term ensures the sub-effects in the target image and the source example are similarly distributed, which is the basis of our assumption in Sec.~\ref{sec:dist-scale} that the posterior probabilities $\mathcal{P}(\ell|x)$ in $T'$ and $S'$ are the same. Fig.~\ref{fig:ana-dist} shows the effects of the distribution term on the \textit{flame} style.
Without distribution constraints, the flames appear randomly in the black background. The distribution term adjusts the flames to better match their spatial distribution as that in the source example.

\textbf{Psycho-Visual Term.} The effects of our psycho-visual term are shown in Fig.~\ref{fig:rep}. The \textit{lava} textures synthesized without the psycho-visual penalty (Fig.~\ref{fig:rep}(a)) densely repeat the red cracks in several regions, which causes obvious unnaturalness. By increasing the penalty, the reuse of the same source textures is greatly restrained (Fig.~\ref{fig:rep}(b)) and our method tends to agilely combine different source patches to create brand-new textures (Fig.~\ref{fig:rep}(c)). Thus, the psycho-visual term can effectively penalize texture over-repetitiveness and encourage new texture creation.

\textbf{Combination of the Three Term.} It is worth noting that \emph{the proposed three terms are complementary}: First, the appearance and distribution terms emphasize local texture patterns and global text sub-effects distributions, respectively. The former depicts low-level color features while the latter exploits complementary mid-level position features. Second, the appearance and distribution terms jointly evaluate objective patch similarities. Meanwhile, the psycho-visual term complements these two terms by incorporating aesthetic subjective evaluations.

\begin{figure*}[t]
\centering
\subfigure{
\includegraphics[width=0.13\linewidth]{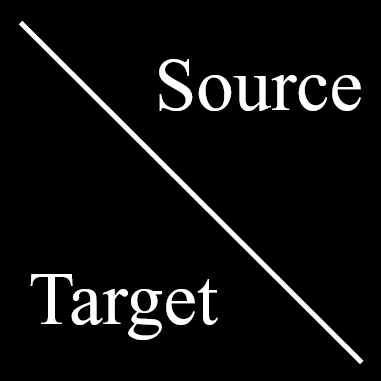}}
\subfigure{
\includegraphics[width=0.13\linewidth]{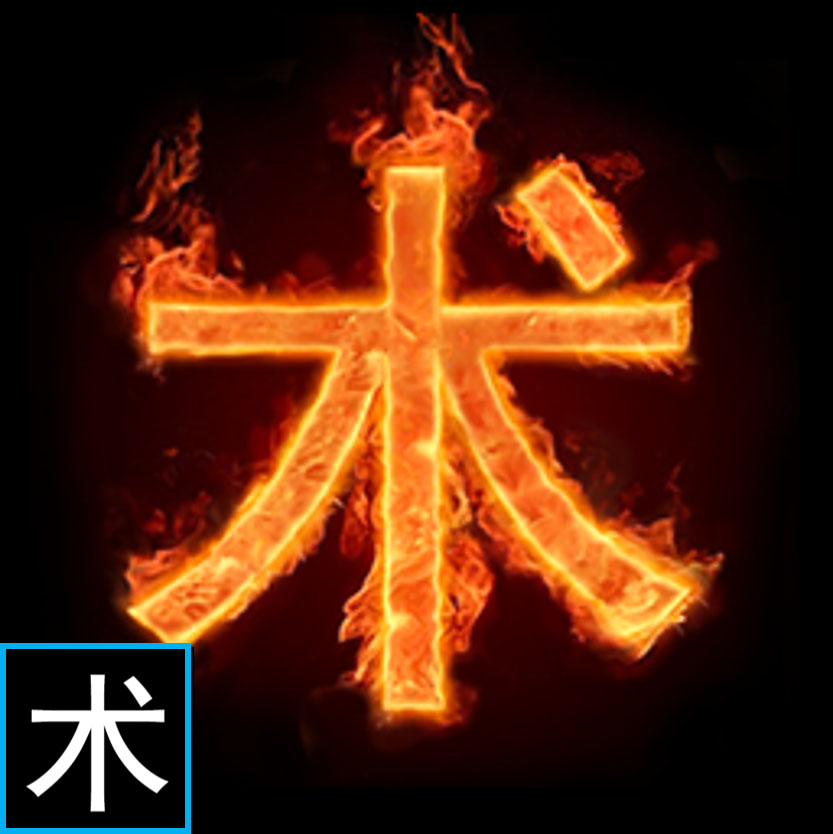}}
\subfigure{
\includegraphics[width=0.13\linewidth]{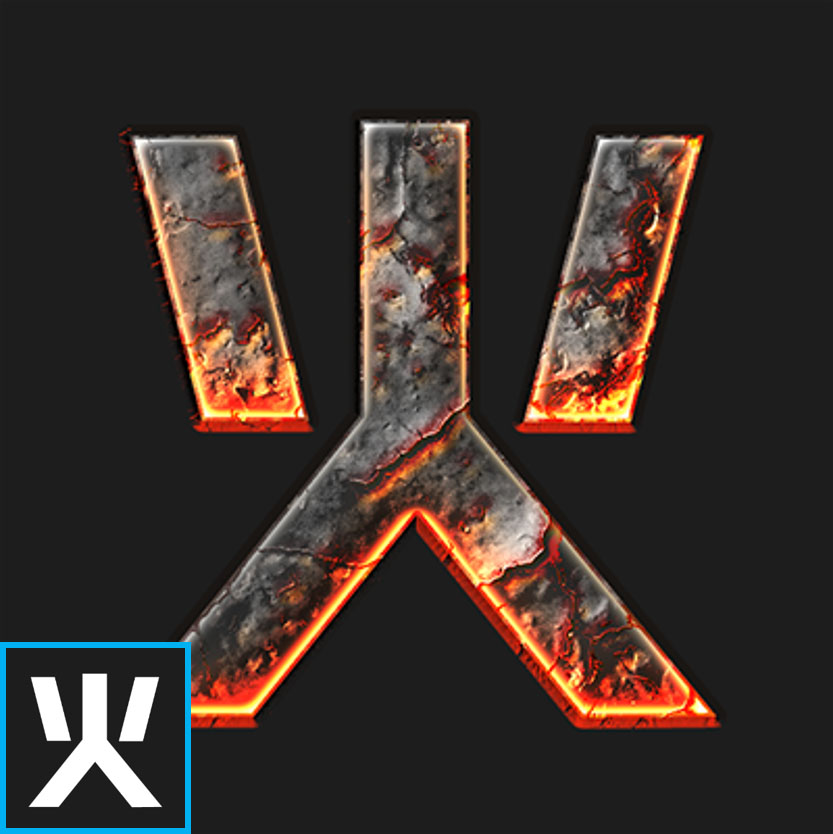}}
\subfigure{
\includegraphics[width=0.13\linewidth]{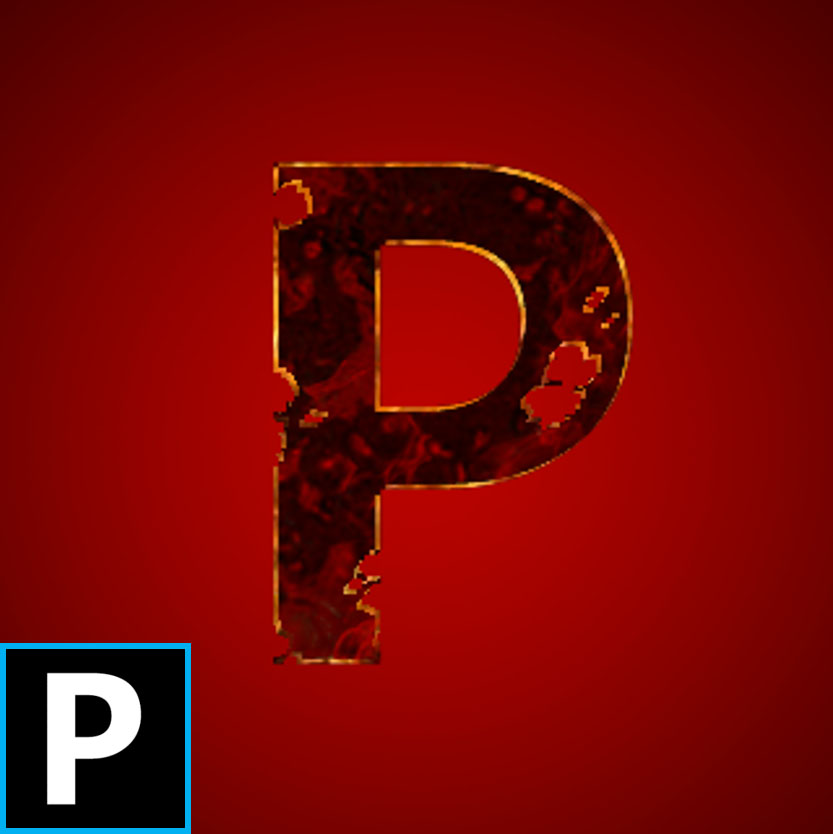}}
\subfigure{
\includegraphics[width=0.13\linewidth]{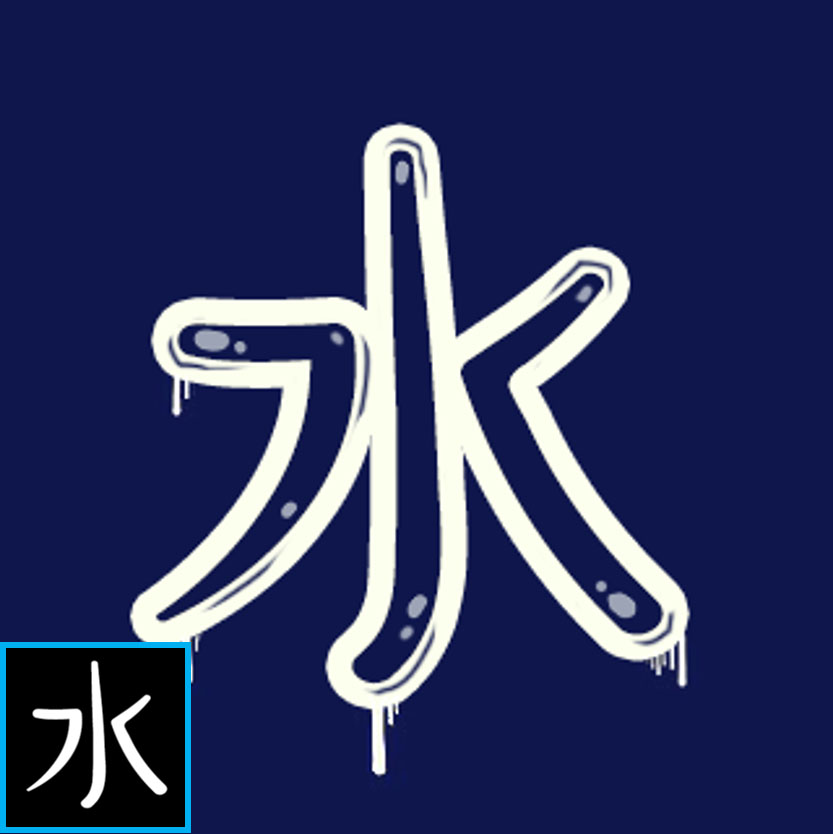}}
\subfigure{
\includegraphics[width=0.13\linewidth]{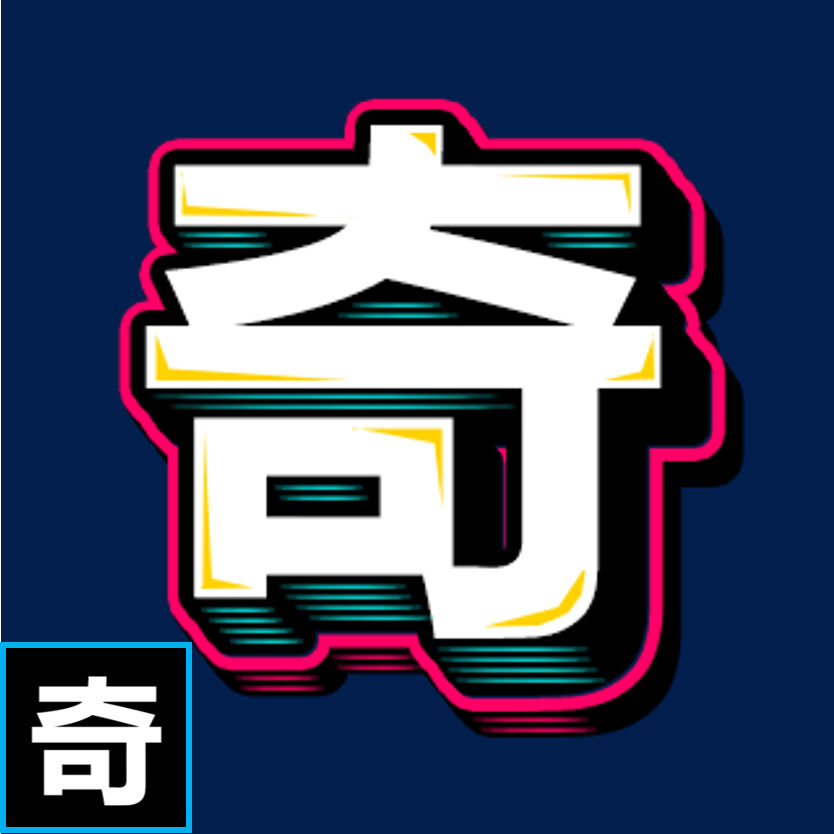}}
\subfigure{
\includegraphics[width=0.13\linewidth]{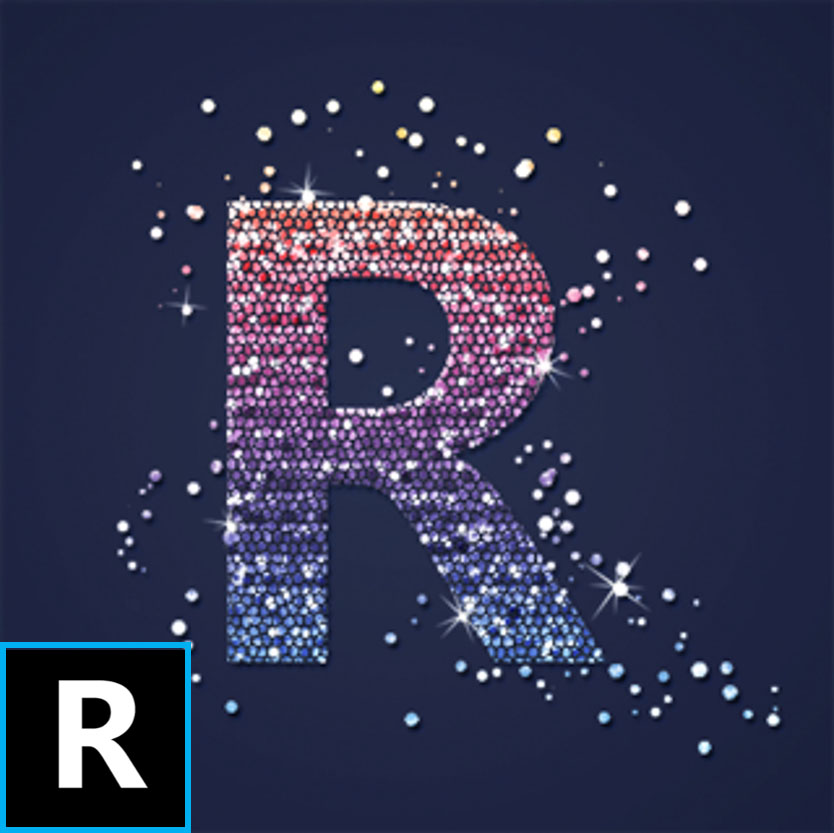}}\vspace{-2mm}
\subfigure{
\includegraphics[width=0.13\linewidth]{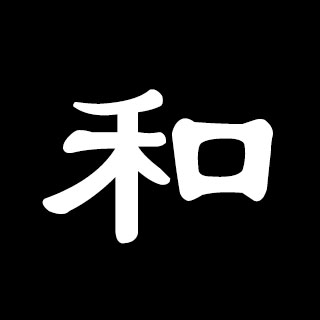}}
\subfigure{
\includegraphics[width=0.13\linewidth]{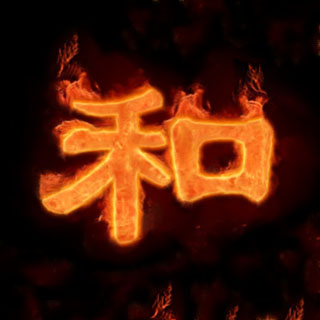}}
\subfigure{
\includegraphics[width=0.13\linewidth]{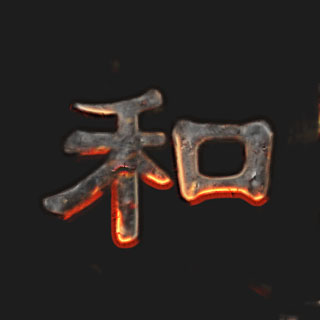}}
\subfigure{
\includegraphics[width=0.13\linewidth]{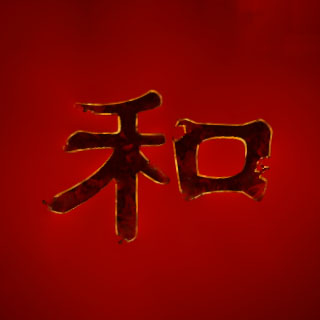}}
\subfigure{
\includegraphics[width=0.13\linewidth]{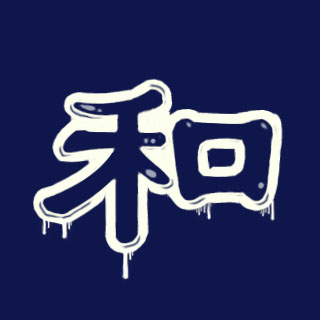}}
\subfigure{
\includegraphics[width=0.13\linewidth]{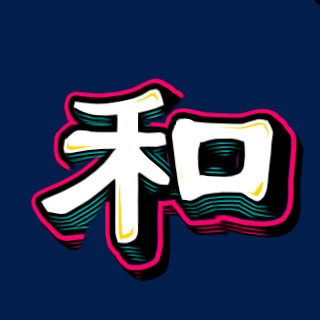}}
\subfigure{
\includegraphics[width=0.13\linewidth]{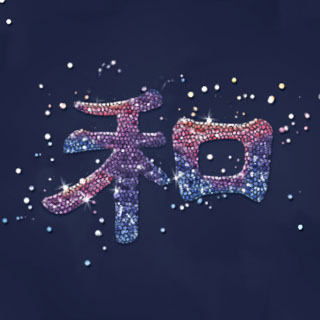}}\vspace{-2mm}
\subfigure{
\includegraphics[width=0.13\linewidth]{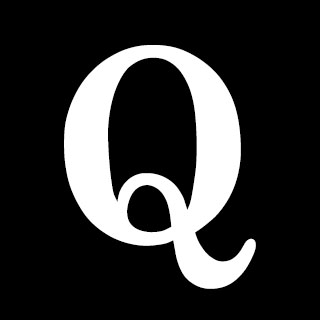}}
\subfigure{
\includegraphics[width=0.13\linewidth]{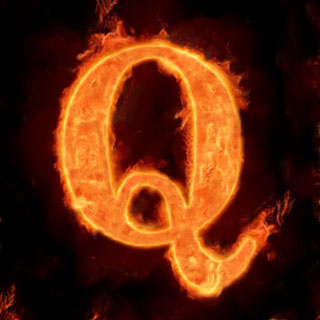}}
\subfigure{
\includegraphics[width=0.13\linewidth]{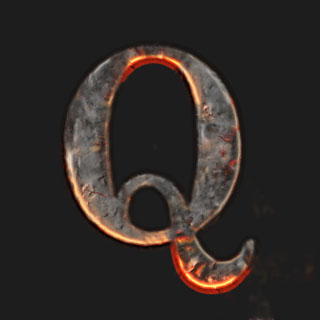}}
\subfigure{
\includegraphics[width=0.13\linewidth]{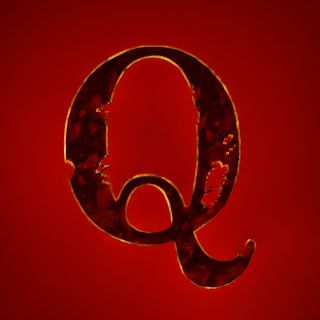}}
\subfigure{
\includegraphics[width=0.13\linewidth]{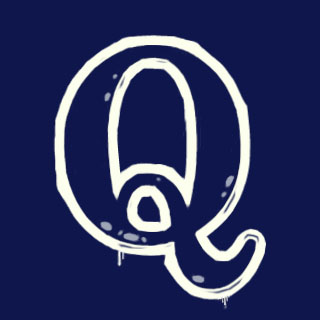}}
\subfigure{
\includegraphics[width=0.13\linewidth]{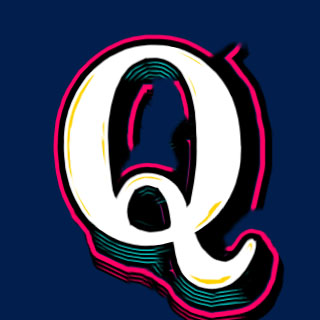}}
\subfigure{
\includegraphics[width=0.13\linewidth]{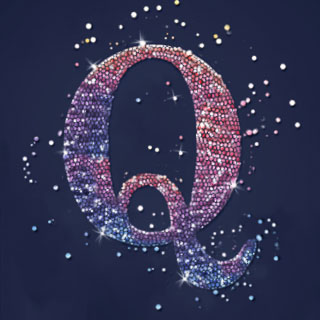}}\addtocounter{subfigure}{-21}\vspace{-2mm}
\subfigure[Target text]{
\includegraphics[width=0.13\linewidth]{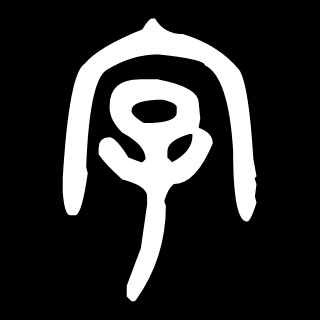}}
\subfigure[\textit{flame}]{
\includegraphics[width=0.13\linewidth]{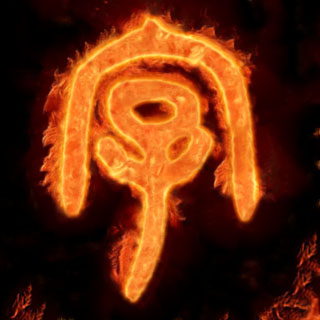}}
\subfigure[\textit{lava}]{
\includegraphics[width=0.13\linewidth]{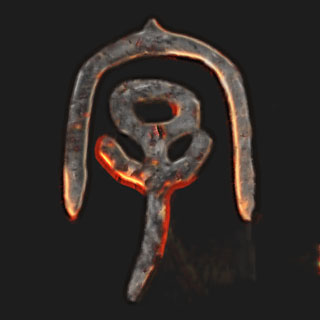}}
\subfigure[\textit{rust}]{
\includegraphics[width=0.13\linewidth]{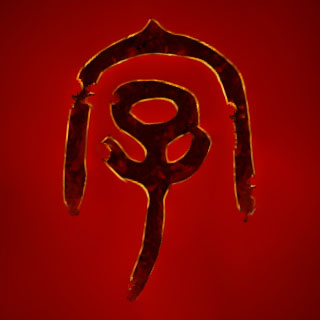}}
\subfigure[\textit{drop}]{
\includegraphics[width=0.13\linewidth]{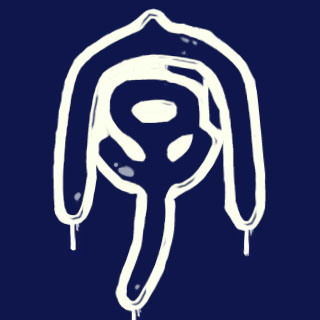}}
\subfigure[\textit{pop}]{
\includegraphics[width=0.13\linewidth]{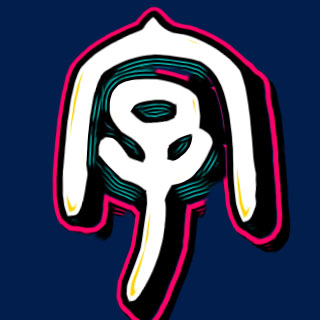}}
\subfigure[\textit{blink}]{
\includegraphics[width=0.13\linewidth]{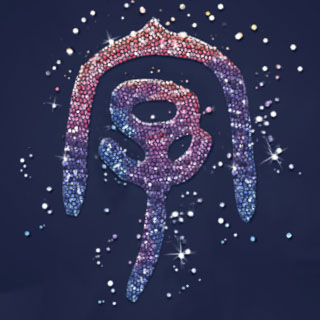}}
\caption{Apply different text effects to representative characters (Chinese, alphabetic, handwriting).}
\label{fig:more}
  \vspace{-1mm}
\end{figure*}

\begin{figure*}
  \centering
  \includegraphics[width=0.96\linewidth]{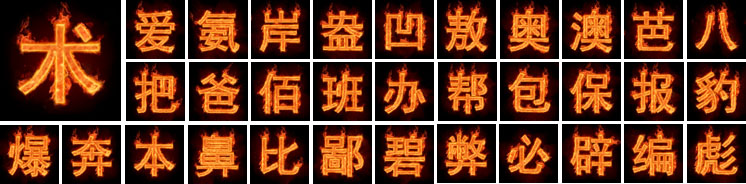}
  \caption{An overview of our \textit{flame} typography library. The bigger image at the top left corner serves as the example to generate the other $774$ characters. The whole library as well as the other stylized libraries can be found in the supplementary material.}\label{fig:library1}
  \vspace{-2mm}
\end{figure*}

\section{Experimental Results}
\label{sec:experiment}
\vspace{-1mm}
In the experiment, the patch size is $5\times5$ and the max scale $L=5$. We build an image pyramid of $10$ levels with a fixed coarsest size ($32\times32$). At level $\ell$, joint patches over scales from $\ell$ to $min(10,\ell+L-1)$ are used. The weights $\lambda_1$, $\lambda_2$ and $\lambda_3$ to balance different terms are set to $0.01$, $0.005$ and $10$, respectively. The parameter $\omega$ for the filter criterion is $0.3$. In addition to the examples in this paper, all of our results and comparisons are included in the supplementary material.

In Fig.~\ref{fig:comparison}, we present a comparison of our algorithm with three state-of-the-art style transfer techniques as well as our baseline.
The first method is the pioneering Image Analogies \cite{Hertzmann2001Image}.
%In their results, the textures are transferred at some point but in a locally repeated and globally random way. And patch boundaries are too evident.
The textures in their results repeat locally and look disordered globally with evident patch boundaries.
The second method is our implementation of Split and Match \cite{Frigo2016Split}, which synthesizes textures using adaptive patch sizes. The original method directly transfers the style in $S'$ to $T$ without the help of $S$. To make a fair comparison, we incorporate the guidance by using $S$ instead of $S'$ in the split stage. This method fails to generate textures in the background and produces plain stylized results.
The third method, Neural Doodle \cite{Champandard2016Semantic}, is based on the combination of MRF and CNN \cite{Li2016Combining} and incorporates semantic maps for analogy guidance. While the color palette of the example text effects is transferred, fine textures are poorly synthesized. The text shape is lost as well.
Our baseline transfers fine textures %and works well on the simple text effects of \textit{silver}.
but fails to keep the overall sub-effects distribution and generates artifacts in the background.
By comparison, the proposed method outperforms state-of-the-art methods, preserving both local textures and the global sub-effects distribution.

In Fig.~\ref{fig:more}, we present an illustration of style transfer from six very different text effects to three representative characters (Chinese, alphabetic, handwriting). This experiment covers challenging transformations between styles, languages and fonts. Thanks to distance normalization and multi-scale strategy, our algorithm accomplishes to transfer the text effects regardless of character shapes and texture scales, providing a solid tool for artistic typography.

Finally, we show our \textit{flame} typography library including as much as $775$ frequently used Chinese characters. Due to the space limitation, only the first $32$ of them are presented in Fig.~\ref{fig:library1}. The whole library as well as the other typography libraries are included in our supplementary material. The extensive synthesis results demonstrate the robustness of our method to varied character shapes.

%-------------------------------------------------------------------------
\section{Conclusion}
\vspace{-2mm}
In this paper, we raise the text effects transfer problem and propose a novel statistics-based method to solve it. We convert the high correlation between the sub-effects patterns and their relative spatial distribution
to the text skeletons into soft constraints for text effects generation. An objective function with three complementary terms is proposed to jointly consider the local multi-scale texture, global sub-effects distribution and visual naturalness. We validate the effectiveness and robustness of our method by comparisons with state-of-the-art style transfer algorithms and extensive artistic typography generations. Future work will concentrate on the composition of the stylized texts and the background photos.

{\small
\bibliographystyle{ieee}
\bibliography{egbib}
}

\clearpage

\begin{figure}[H]
\centering
\subfigure[Source raw text $S$]{
\includegraphics[width=0.32\linewidth]{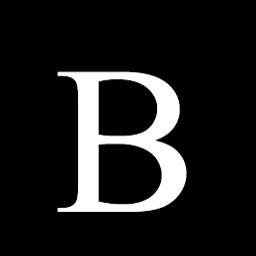}}
\subfigure[Source text effects $S'$]{
\includegraphics[width=0.32\linewidth]{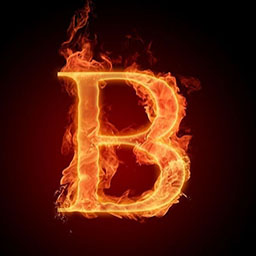}}
\subfigure[Target raw text $T$]{
\includegraphics[width=0.32\linewidth]{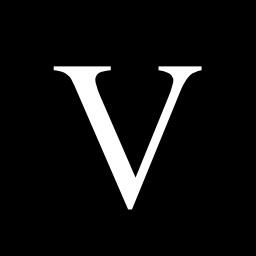}}\\
\subfigure[Image Analogies~\protect\cite{Hertzmann2001Image}]{
\includegraphics[width=0.32\linewidth]{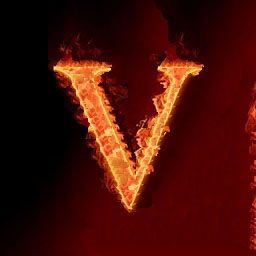}}
\subfigure[Split and Match~\protect\cite{Frigo2016Split}]{
\includegraphics[width=0.32\linewidth]{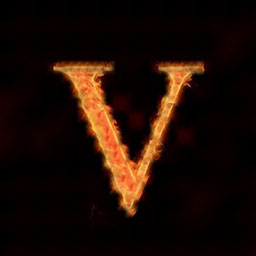}}
\subfigure[Neural Doodles~\protect\cite{Champandard2016Semantic}]{
\includegraphics[width=0.32\linewidth]{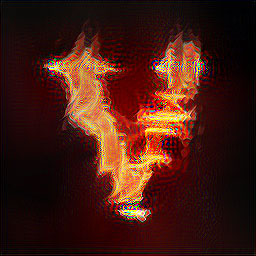}}\\
\subfigure[PatchTable~\protect\cite{barnes2015}]{
\includegraphics[width=0.32\linewidth]{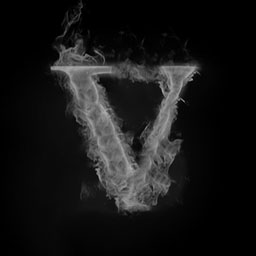}}
\subfigure[Baseline]{
\includegraphics[width=0.32\linewidth]{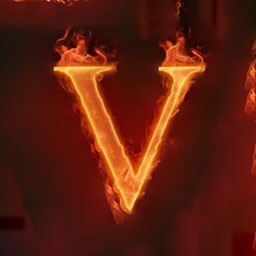}}
\subfigure[Proposed method]{
\includegraphics[width=0.32\linewidth]{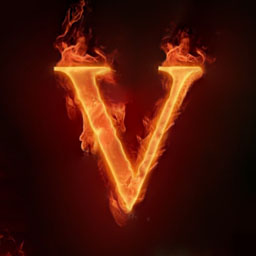}}
\caption{Comparison with state-of-the-art methods on the text effects of \textit{flame}\protect\footnotemark[4]. (a) Input source raw text. (b) Input source text effects. (c) Target text. (d) Results of Image Analogies~\cite{Hertzmann2001Image}. (e) Results of Split and Match~\cite{Frigo2016Split}.  (f) Results of Neural Doodles~\cite{Champandard2016Semantic}. (g) Results of PatchTable~\cite{barnes2015} (h) Results of our baseline method. (i) Results of the proposed method.}
\label{fig:comparison-flame}
\end{figure}

\begin{figure}[H]
\centering
\subfigure[Source raw text $S$]{
\includegraphics[width=0.32\linewidth]{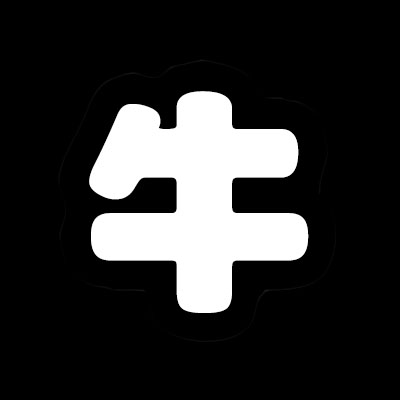}}
\subfigure[Source text effects $S'$]{
\includegraphics[width=0.32\linewidth]{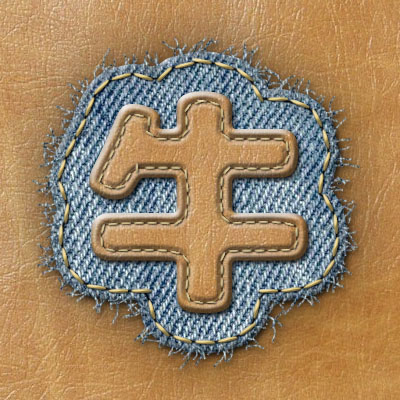}}
\subfigure[Target raw text $T$]{
\includegraphics[width=0.32\linewidth]{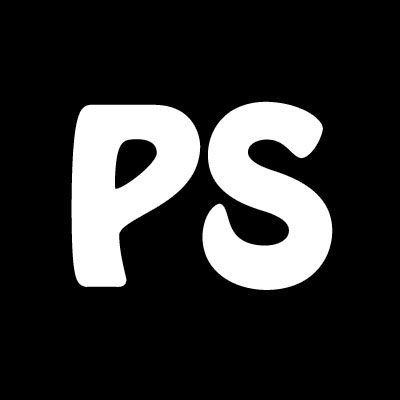}}\\
\subfigure[Image Analogies~\protect\cite{Hertzmann2001Image}]{
\includegraphics[width=0.32\linewidth]{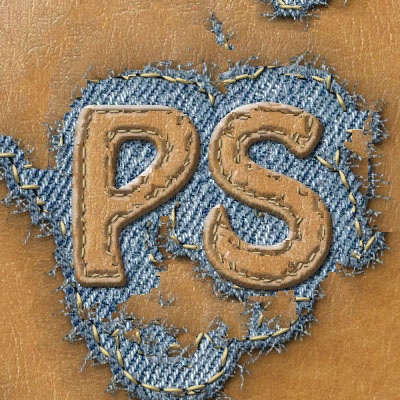}}
\subfigure[Split and Match~\protect\cite{Frigo2016Split}]{
\includegraphics[width=0.32\linewidth]{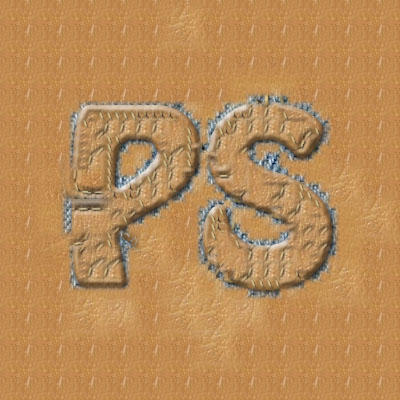}}
\subfigure[Neural Doodles~\protect\cite{Champandard2016Semantic}]{
\includegraphics[width=0.32\linewidth]{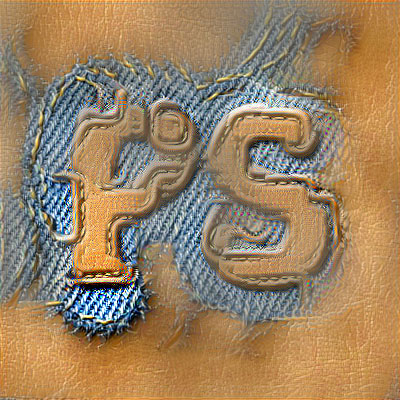}}\\
\subfigure[PatchTable~\protect\cite{barnes2015}]{
\includegraphics[width=0.32\linewidth]{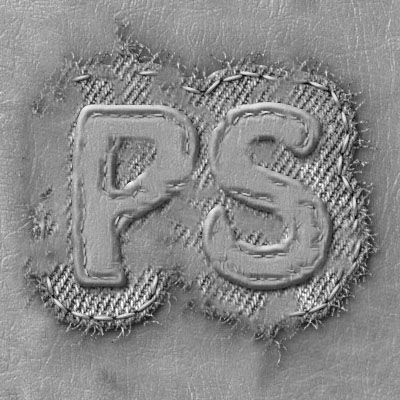}}
\subfigure[Baseline]{
\includegraphics[width=0.32\linewidth]{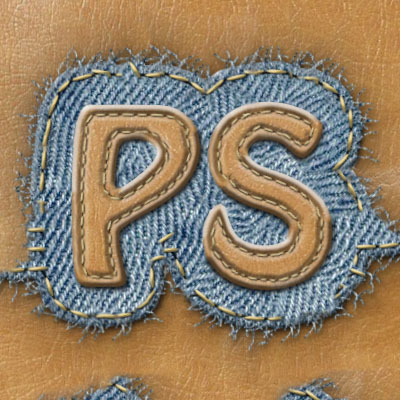}}
\subfigure[Proposed method]{
\includegraphics[width=0.32\linewidth]{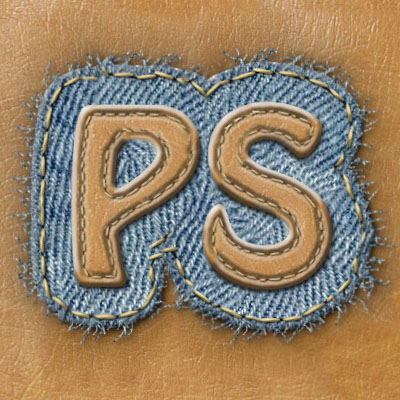}}
\caption{Comparison with state-of-the-art methods on the text effects of \textit{denim fabric}\protect\footnotemark[5]. (a) Input source raw text. (b) Input source text effects. (c) Target text. (d) Results of Image Analogies~\cite{Hertzmann2001Image}. (e) Results of Split and Match~\cite{Frigo2016Split}.  (f) Results of Neural Doodles~\cite{Champandard2016Semantic}. (g) Results of PatchTable~\cite{barnes2015} (h) Results of our baseline method. (i) Results of the proposed method.}
\label{fig:comparison-jeans}
\end{figure}

\begin{figure}[H]
\centering
\subfigure[Source raw text $S$]{
\includegraphics[width=0.32\linewidth]{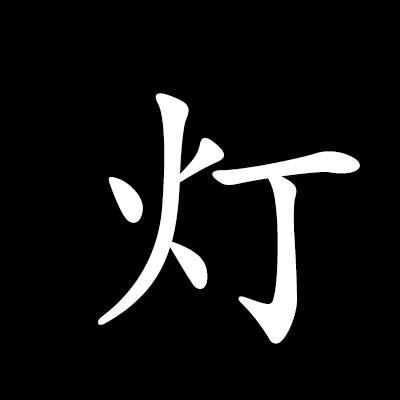}}
\subfigure[Source text effects $S'$]{
\includegraphics[width=0.32\linewidth]{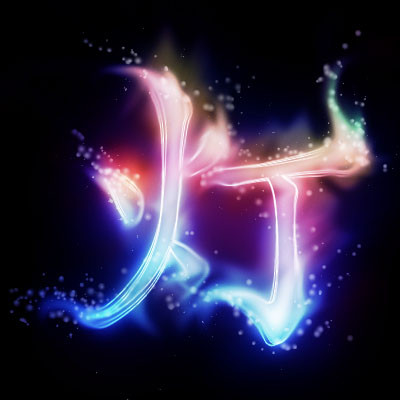}}
\subfigure[Target raw text $T$]{
\includegraphics[width=0.32\linewidth]{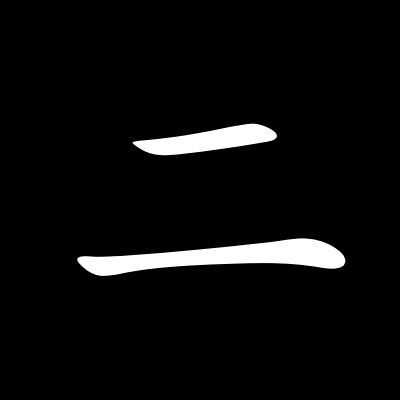}}\\
\subfigure[Image Analogies~\protect\cite{Hertzmann2001Image}]{
\includegraphics[width=0.32\linewidth]{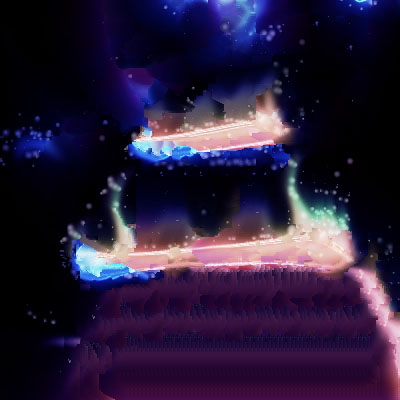}}
\subfigure[Split and Match~\protect\cite{Frigo2016Split}]{
\includegraphics[width=0.32\linewidth]{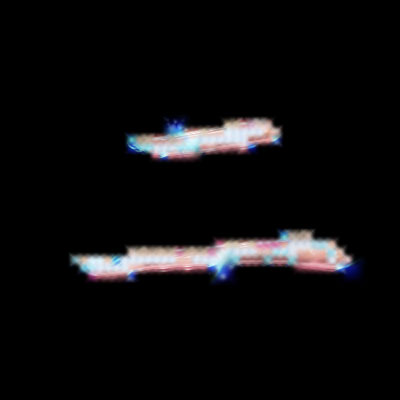}}
\subfigure[Neural Doodles~\protect\cite{Champandard2016Semantic}]{
\includegraphics[width=0.32\linewidth]{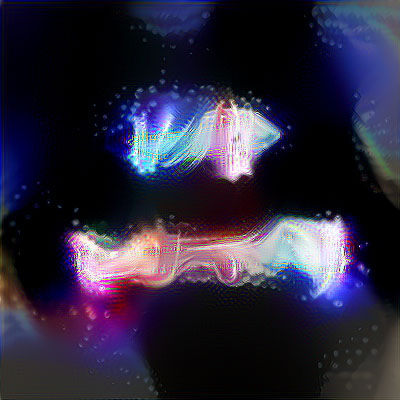}}\\
\subfigure[PatchTable~\protect\cite{barnes2015}]{
\includegraphics[width=0.32\linewidth]{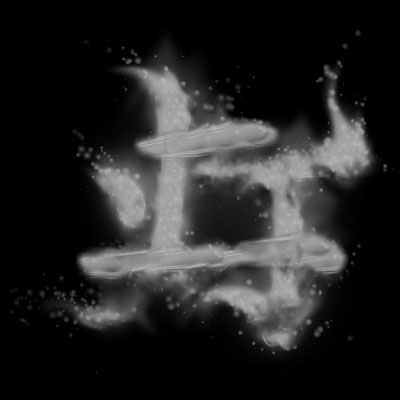}}
\subfigure[Baseline]{
\includegraphics[width=0.32\linewidth]{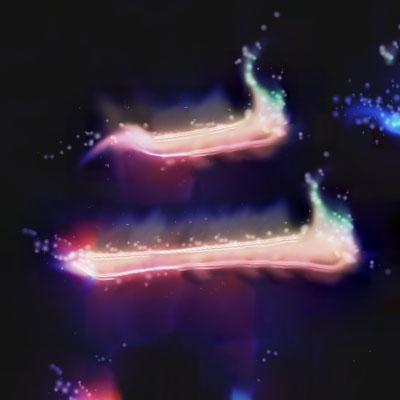}}
\subfigure[Proposed method]{
\includegraphics[width=0.32\linewidth]{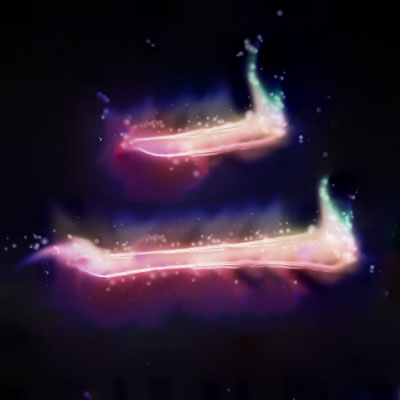}}
\caption{Comparison with state-of-the-art methods on the text effects of \textit{neon}\protect\footnotemark[6]. (a) Input source raw text. (b) Input source text effects. (c) Target text. (d) Results of Image Analogies~\cite{Hertzmann2001Image}. (e) Results of Split and Match~\cite{Frigo2016Split}.  (f) Results of Neural Doodles~\cite{Champandard2016Semantic}. (g) Results of PatchTable~\cite{barnes2015} (h) Results of our baseline method. (i) Results of the proposed method.}
\label{fig:comparison-neon}
\end{figure}

\begin{figure}[H]
\centering
\subfigure[\textit{water}]{
\includegraphics[width=0.32\linewidth]{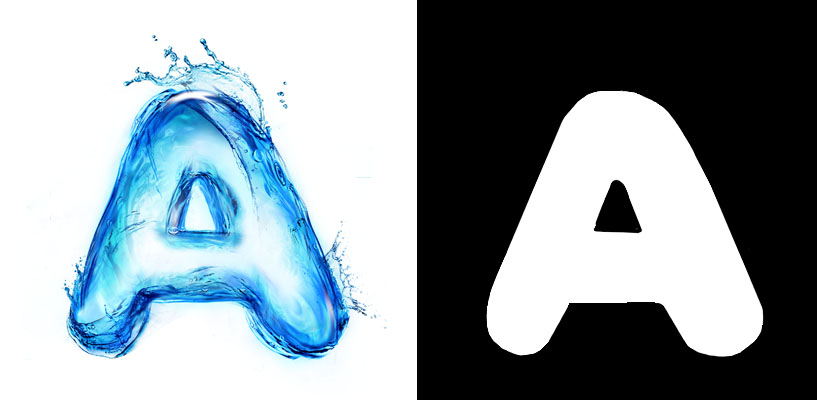}}
\subfigure[\textit{flame}]{
\includegraphics[width=0.32\linewidth]{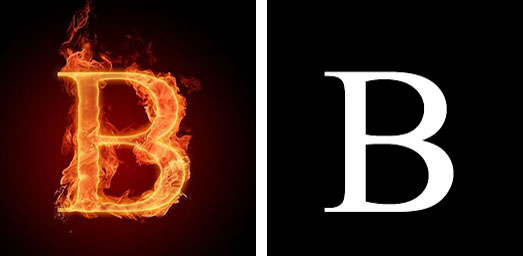}}
\subfigure[\textit{rust}]{
\includegraphics[width=0.32\linewidth]{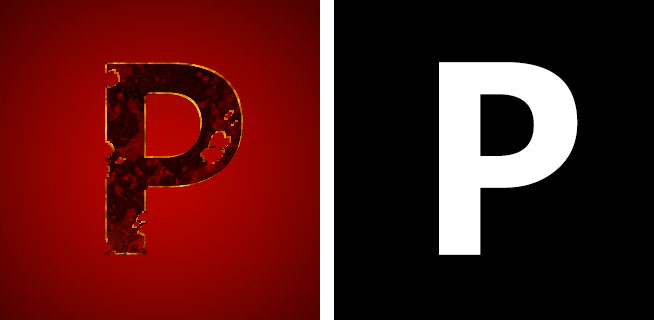}}
\subfigure[\textit{blink}]{
\includegraphics[width=0.32\linewidth]{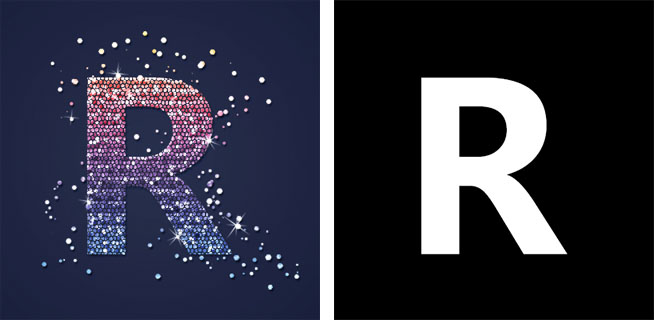}}
\subfigure[\textit{drop}]{
\includegraphics[width=0.32\linewidth]{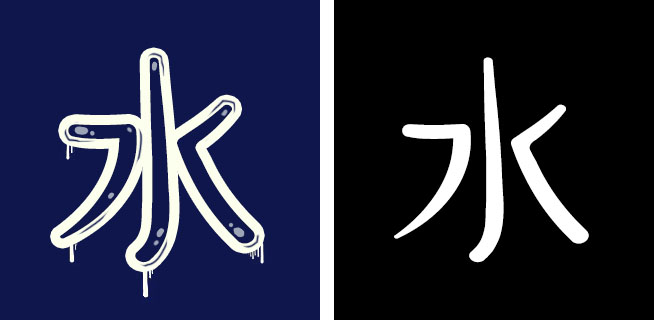}}
\subfigure[\textit{silver}]{
\includegraphics[width=0.32\linewidth]{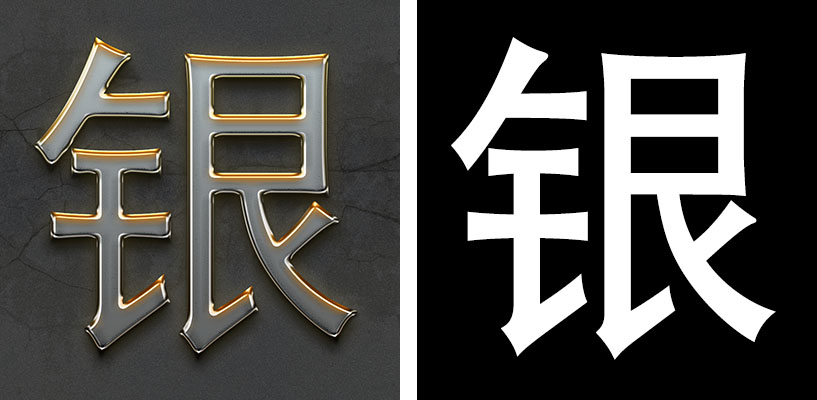}}
\caption{Six source text effects images\protect\footnotemark[7]~for experiments. Text effects: \textit{water}, \textit{flame}, \textit{rust}, \textit{blink}, \textit{drop}, \textit{silver}.}
\label{fig:source}
\vspace{-4mm}
\end{figure}

\begin{figure}[H]
\centering
\subfigure[Chinese character \textit{Chen}]{
\includegraphics[width=0.235\linewidth]{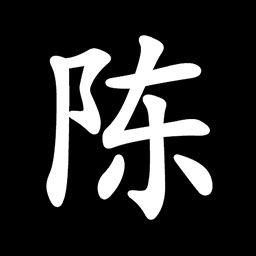}}
\subfigure[Chinese character \textit{He}]{
\includegraphics[width=0.235\linewidth]{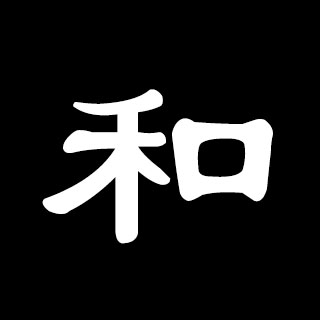}}
\subfigure[Alphabetic character \textit{Q}]{
\includegraphics[width=0.235\linewidth]{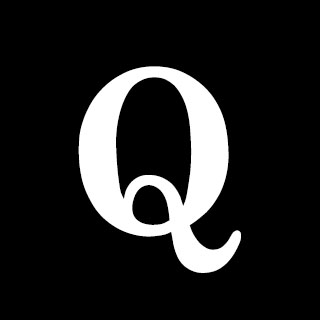}}
\subfigure[Chinese character \textit{Yuan}]{
\includegraphics[width=0.235\linewidth]{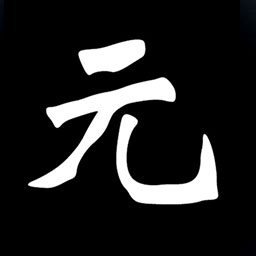}}
\subfigure[Handwriting character \textit{Zi}]{
\includegraphics[width=0.235\linewidth]{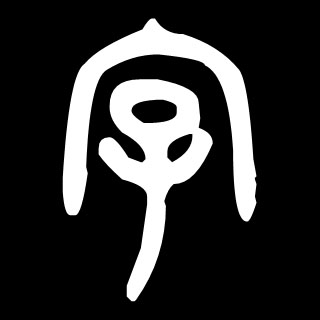}}
\subfigure[Alphabetic character \textit{R}]{
\includegraphics[width=0.235\linewidth]{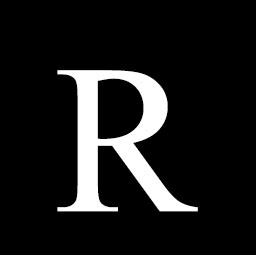}}
\subfigure[Chinese character \textit{Ai}]{
\includegraphics[width=0.235\linewidth]{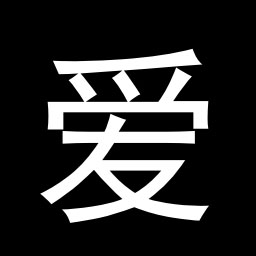}}
\subfigure[Handwriting character \textit{Qi}]{
\includegraphics[width=0.235\linewidth]{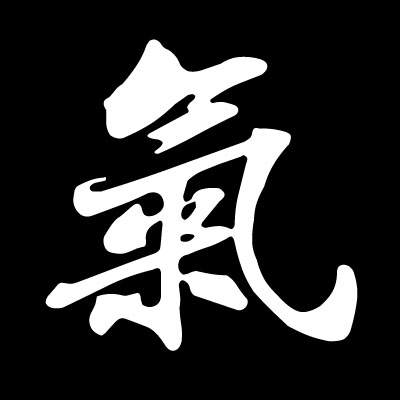}}
\caption{Eight target representative characters (Chinese, alphabetic, handwriting) for experiments. Target texts: \textit{Chen}, \textit{He}, \textit{Q}, \textit{Yuan}, \textit{Zi}, \textit{R}, \textit{Ai}, \textit{Qi}.}
\label{fig:target}
\end{figure}

\begin{figure}[H]
\centering
\subfigure[Overview of our \textit{flame} typography library with $62$ alphabetic characters and Arabic numerals]{
\includegraphics[width=0.96\linewidth]{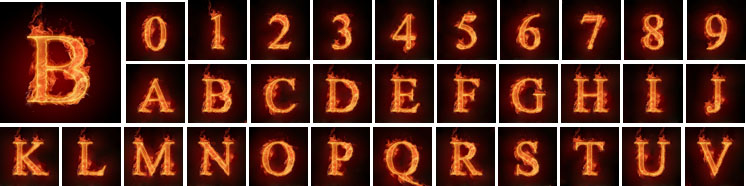}}
\subfigure[Overview of our \textit{smoke} typography library with $62$ alphabetic and Arabic numerals characters]{
\includegraphics[width=0.96\linewidth]{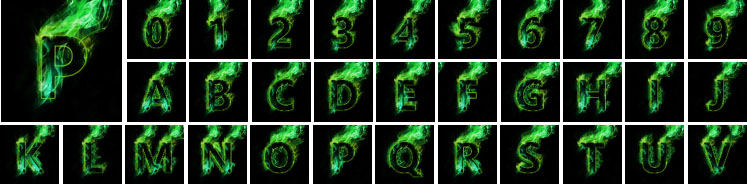}}
\subfigure[Overview of our \textit{flame} typography library with $775$ frequently used Chinese characters]{
\includegraphics[width=0.96\linewidth]{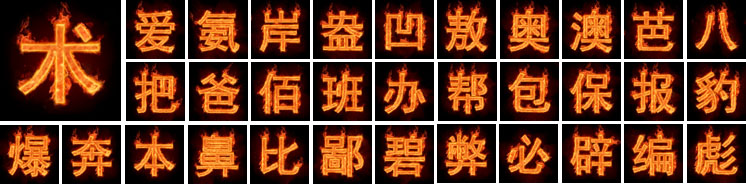}}
\subfigure[Overview of our \textit{neon} typography library with $775$ frequently used Chinese characters]{
\includegraphics[width=0.96\linewidth]{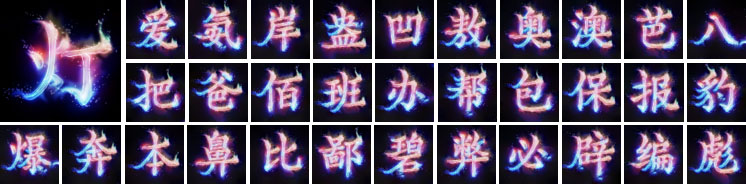}}
\caption{An overview of our \textit{flame}, \textit{smoke} and \textit{neon} typography libraries. The bigger image at the top left corner serves as the example to generate the other characters. Due to the space limitation, only the first $32$ characters of each library are presented.}
\label{fig:lib}
\end{figure}

\onecolumn
\begin{figure*}
\vspace{32mm}
  \centering
  \includegraphics[width=0.96\linewidth]{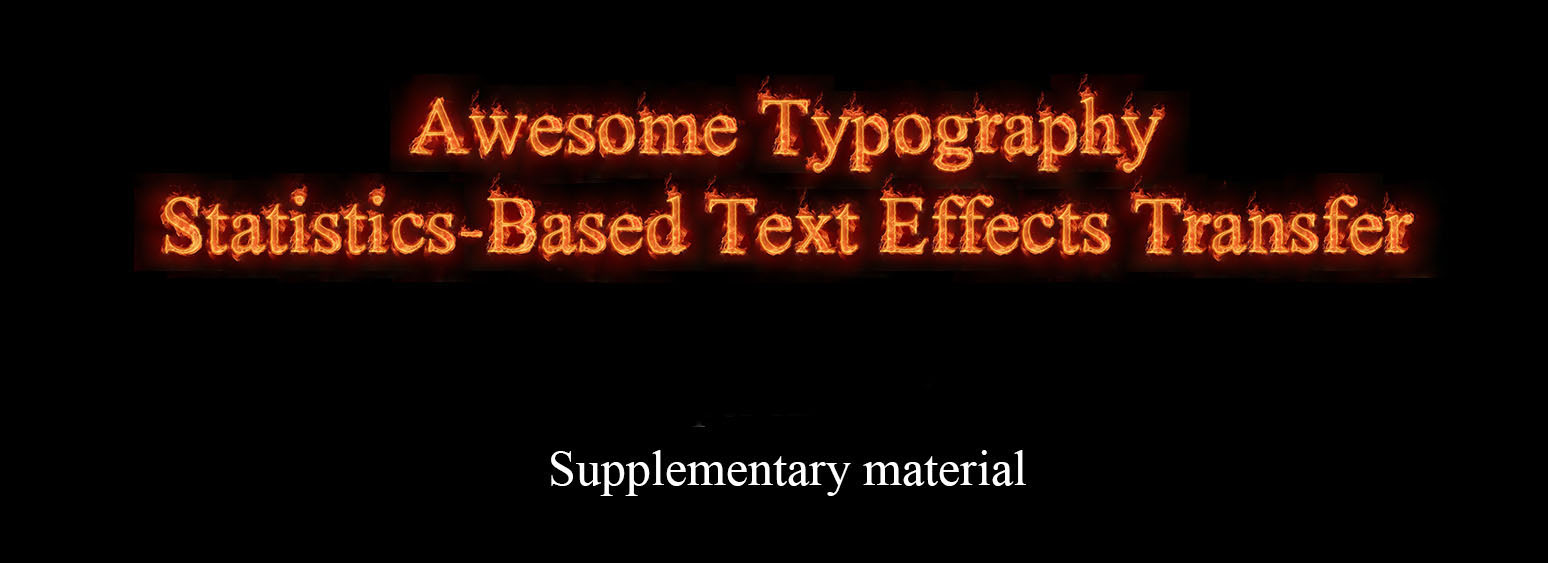}
\end{figure*}
\vspace{2mm}
As supplementary material of our paper, we present the following contents:
\vspace{2mm}
\begin{itemize}
  \item Correlations between patch patterns and different modes in $30$ text effects images. (Table \ref{tb:2})\vspace{2mm}
  \item Comparison of our text effects transfer approach with state-of-the-art methods. (Figs. \ref{fig:comparison-smoke}$-$\ref{fig:comparison-neon})\vspace{2mm}
  \item Illustration of style transfer from various text effects to different characters. (Figs. \ref{fig:source}$-$\ref{fig:resultD})\vspace{2mm}
  \item An overview of our four artistic typography libraries. (Fig.~\ref{fig:lib})
\end{itemize}

\clearpage

\section{Correlations Between Patch Patterns and Different Modes}

\begin{table} [htb]
\caption{Correlations between patch patterns and different modes.}
\label{tb:2}
\centering
\begin{tabular}{c|ccccc|ccccc}
\hline
\multirow{2}{*}{Images\footnotemark[1]} & \multicolumn{5}{c|}{color} &  \multicolumn{5}{c}{scale} \\
\cline{2-11}
~ &  rand & gird & angle & ring & dist & rand & gird & angle & ring & dist\\
\hline
01 & 0.061  & 0.115  & 0.120  & 0.105  & \textbf{0.167}  & 0.134  & \textbf{0.833}  & 0.462  & 0.571  & 0.394 \\
02 & 0.061  & 0.088  & 0.100  & 0.088  & \textbf{0.139}  & 0.178  & 1.219  & 0.241  & \textbf{1.488}  & 1.474 \\
03 & 0.064  & 0.104  & \textbf{0.139}  & 0.090  & 0.125  & 0.131  & 0.872  & 0.496  & 0.650  & \textbf{1.208} \\
04 & 0.067  & 0.140  & 0.109  & 0.133  & \textbf{0.197}  & 0.175  & 0.418  & 0.327  & 0.567  & \textbf{0.606} \\
05 & 0.059  & 0.123  & 0.115  & 0.125  & \textbf{0.152}  & 0.158  & 0.471  & 0.232  & 0.593  & \textbf{0.671} \\
06 & 0.059  & 0.095  & 0.101  & 0.088  & \textbf{0.133}  & 0.193  & \textbf{1.053}  & 0.721  & 0.406  & 1.021 \\
07 & 0.063  & 0.126  & 0.115  & 0.130  & \textbf{0.147}  & 0.148  & 0.434  & 0.222  & 0.463  & \textbf{1.088} \\
08 & 0.068  & 0.116  & 0.125  & 0.111  & \textbf{0.134}  & 0.144  & 1.109  & 0.492  & 1.014  & \textbf{1.113} \\
09 & 0.062  & 0.106  & 0.100  & 0.111  & \textbf{0.144}  & 0.154  & 0.916  & 0.426  & 0.804  & \textbf{1.304} \\
10 & 0.063  & 0.114  & 0.109  & 0.109  & \textbf{0.137}  & 0.160  & 0.536  & 0.291  & 0.671  & \textbf{1.027} \\
11 & 0.064  & 0.116  & 0.115  & 0.121  & \textbf{0.136}  & 0.163  & 0.385  & 0.267  & 0.539  & \textbf{0.817} \\
12 & 0.064  & 0.123  & 0.111  & 0.115  & \textbf{0.168}  & 0.171  & 0.584  & 0.332  & 0.725  & \textbf{0.867} \\
13 & 0.063  & 0.096  & 0.105  & 0.122  & \textbf{0.131}  & 0.113  & \textbf{1.099}  & 0.304  & 0.972  & 0.766 \\
14 & 0.062  & 0.097  & 0.112  & 0.107  & \textbf{0.147}  & 0.164  & \textbf{0.765}  & 0.518  & 0.479  & 0.672 \\
15 & 0.061  & 0.099  & 0.118  & 0.102  & \textbf{0.144}  & 0.162  & \textbf{0.851}  & 0.403  & 0.797  & 0.547 \\
16 & 0.065  & 0.104  & 0.129  & 0.104  & \textbf{0.193}  & 0.110  & 1.143  & 0.630  & 0.551  & \textbf{1.568} \\
17 & 0.063  & 0.086  & 0.125  & 0.095  & \textbf{0.137}  & 0.175  & 1.078  & 0.868  & 0.488  & \textbf{1.303} \\
18 & 0.065  & 0.123  & 0.127  & 0.117  & \textbf{0.132}  & 0.145  & \textbf{0.546}  & 0.282  & 0.514  & 0.505 \\
19 & 0.061  & 0.095  & 0.126  & 0.091  & \textbf{0.142}  & 0.165  & 0.644  & 0.772  & 0.255  & \textbf{0.900} \\
20 & 0.067  & 0.112  & \textbf{0.143}  & 0.120  & 0.142  & 0.156  & \textbf{0.944}  & 0.362  & 0.572  & 0.904 \\
21 & 0.063  & 0.093  & 0.111  & 0.087  & \textbf{0.129}  & 0.163  & \textbf{0.978}  & 0.745  & 0.654  & 0.427 \\
22 & 0.062  & 0.094  & 0.128  & 0.100  & \textbf{0.141}  & 0.110  & 1.245  & 0.885  & 0.466  & \textbf{1.658} \\
23 & 0.065  & 0.120  & 0.138  & 0.109  & \textbf{0.145}  & 0.169  & 0.377  & \textbf{0.377}  & 0.135  & 0.211 \\
24 & 0.064  & 0.097  & 0.115  & 0.095  & \textbf{0.216}  & 0.151  & 0.655  & 0.480  & 0.458  & \textbf{0.783} \\
25 & 0.064  & 0.105  & 0.140  & 0.109  & \textbf{0.147}  & 0.130  & \textbf{0.717}  & 0.671  & 0.455  & 0.661 \\
26 & 0.066  & 0.107  & 0.116  & 0.112  & \textbf{0.141}  & 0.128  & 1.076  & 0.678  & 0.769  & \textbf{1.465} \\
27 & 0.060  & 0.118  & 0.143  & 0.110  & \textbf{0.185}  & 0.132  & 1.144  & 0.752  & 0.475  & \textbf{1.235} \\
28 & 0.058  & 0.111  & 0.128  & 0.096  & \textbf{0.128}  & 0.175  & 0.934  & 0.683  & 0.491  & \textbf{1.369} \\
29 & 0.061  & 0.087  & 0.108  & 0.082  & \textbf{0.136}  & 0.124  & 0.290  & 0.306  & 0.330  & \textbf{1.165} \\
30 & 0.063  & 0.077  & \textbf{0.111}  & 0.077  & 0.107  & 0.205  & 0.464  & 0.353  & 0.354  & \textbf{0.783} \\
\hline
Average & 0.063  & 0.106  & 0.119  & 0.105  & \textbf{0.147}  & 0.153  & 0.793  & 0.486  & 0.590  & \textbf{0.950} \\
\hline
\end{tabular}
\end{table}

\footnotetext[1]{Image credits:\\
01: \url{http://www.zcool.com.cn/work/ZMTg1MTgzMDA=.html}~~
02: \url{http://www.zcool.com.cn/work/ZMTc4MTM5MDQ=.html}\\
03: \url{http://www.zcool.com.cn/work/ZMTc1MzI4MzI=.html}~~
04: \url{http://www.zcool.com.cn/work/ZMTc0NDIxMDg=.html}\\
05: \url{http://www.zcool.com.cn/work/ZMTcwNjEwMTI=.html}~~
06: \url{http://www.zcool.com.cn/work/ZNTE3MTAxMg==.html}\\
07: \url{http://www.zcool.com.cn/work/ZMTU0Mzk3ODg=.html}~~
08: \url{http://www.zcool.com.cn/work/ZMTUxMDc0MDQ=.html}\\
09: \url{http://www.zcool.com.cn/work/ZMTQ0MjY4OTI=.html}~~
10: \url{http://www.zcool.com.cn/work/ZMTQ0MjU3NDA=.html}\\
11: \url{http://www.zcool.com.cn/work/ZNjExMzA0NA==.html}~~
12: \url{http://www.zcool.com.cn/work/ZNjg1MTg1Ng==.html}\\
13: \url{http://www.zcool.com.cn/work/ZMTg5NjM3MzI=/1.html}\\
14: \url{http://www.zcool.com.cn/work/ZMTg5NTg2MDQ=/2.html}\\
15: \url{http://www.zcool.com.cn/work/ZMTg5NDk0NDg=.html}~~
16: \url{http://www.zcool.com.cn/work/ZMTMwMzM5NzI=.html}\\
17: \url{http://www.68ps.com/jc/big_ps_wz.asp?id=3773}~~~~~~~~~
18: \url{http://www.zcool.com.cn/work/ZMTI5MDczMjQ=.html}\\
19: \url{http://www.zcool.com.cn/work/ZNzUzNTkwMA==/2.html}\\
20: \url{http://www.zcool.com.cn/work/ZNzUzNTkwMA==/3.html}\\
21: \url{http://www.zcool.com.cn/work/ZNTE3MTAxMg==.html}~~
22: \url{http://www.zcool.com.cn/work/ZNTEwNDA5Mg==.html}\\
23: \url{http://www.zcool.com.cn/work/ZNDEyMzc0NA==.html}~~
24: \url{http://www.zcool.com.cn/work/ZNzM3NDA5Ng==.html}\\
25: \url{http://www.zcool.com.cn/work/ZNDIzMDMzMg==.html}~~
26: \url{http://www.zcool.com.cn/work/ZMzM3NjQ4NA==.html}\\
27: \url{http://www.zcool.com.cn/work/ZMzM3NjQ4NA==.html}\\
28: \url{http://www.chinaz.com/design/2015/0604/412039.shtml}\\
29: \url{http://www.68ps.com/jc/big_ps_wz.asp?id=3916}~~~~~~~~~
30: \url{http://www.68ps.com/jc/big_ps_wz.asp?id=3854}\\
}

\newpage

\section{Comparisons with State-of-the-Art Methods}
\vspace{-2mm}
\def\fileName {12}
\def\trgName {23}

\begin{figure}[H]
\centering
\subfigure[Source raw text $S$]{
\includegraphics[width=0.32\linewidth]{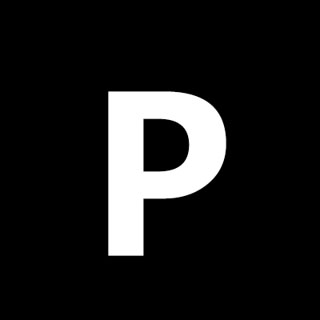}}
\subfigure[Source text effects $S'$]{
\includegraphics[width=0.32\linewidth]{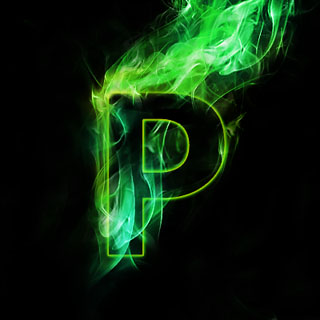}}
\subfigure[Target raw text $T$]{
\includegraphics[width=0.32\linewidth]{supp/\trgName-text.jpg}}\vspace{-1mm}\\
\subfigure[Image Analogies~\protect\cite{Hertzmann2001Image}]{
\includegraphics[width=0.32\linewidth]{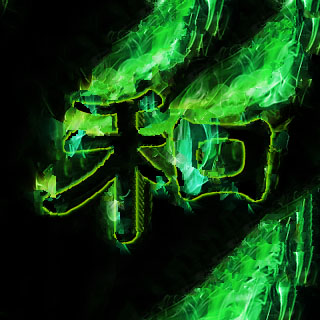}}
\subfigure[Split and Match~\protect\cite{Frigo2016Split}]{
\includegraphics[width=0.32\linewidth]{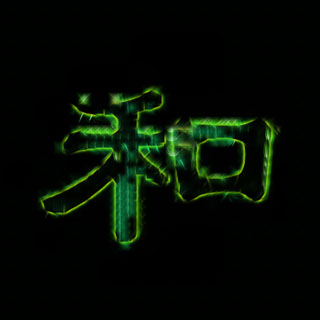}}
\subfigure[Neural Doodles~\protect\cite{Champandard2016Semantic}]{
\includegraphics[width=0.32\linewidth]{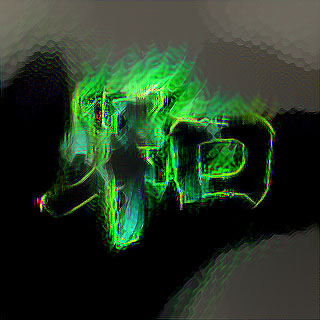}}\vspace{-1mm}\\
\subfigure[PatchTable\protect\cite{barnes2015}]{
\includegraphics[width=0.32\linewidth]{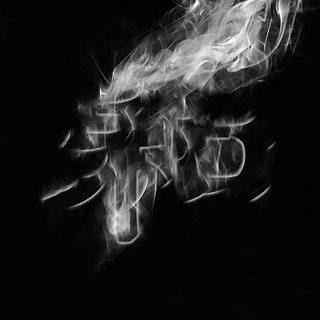}}
\subfigure[Baseline]{
\includegraphics[width=0.32\linewidth]{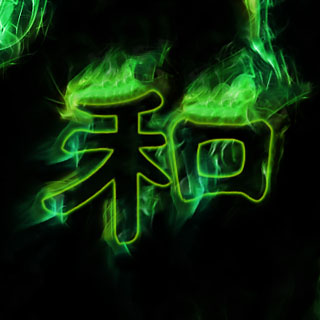}}
\subfigure[Proposed method]{
\includegraphics[width=0.32\linewidth]{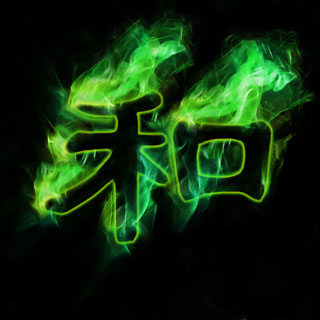}}
\caption{Comparison with state-of-the-art methods on the text effects of \textit{smoke}\protect\footnotemark[2]. (a) Input source raw text. (b) Input source text effects. (c) Target text. (d) Results of Image Analogies~\cite{Hertzmann2001Image}. (e) Results of Split and Match~\cite{Frigo2016Split}.  (f) Results of Neural Doodles~\cite{Champandard2016Semantic}. (g) Results of PatchTable\protect\footnotemark[3]~\cite{barnes2015} (h) Results of our baseline method. (i) Results of the proposed method.}
\label{fig:comparison-smoke}
\end{figure}

\footnotetext[2]{Image credits: Created by us under the design tutorial of~ \url{http://photo.renren.com/photo/249458089/photo-3396512368/v7}}
\footnotetext[3]{PatchTable~\cite{barnes2015} does not transfer colors. Thus, it produces black and white results. }

\def\fileName {4}
\def\trgName {32}

\footnotetext[4]{Image credits:~\url{http://www.phombo.com/wallpapers/fire-letters-wallpapers-hd-3000x3000-a-z0-9/page-1/}}

\def\fileName {10}
\def\trgName {35}

\footnotetext[5]{Image credits: Created by us under the design tutorial of~ \url{http://www.zcool.com.cn/article/ZMTQ1NDA0.html}}

\def\fileName {5}
\def\trgName {122}

\footnotetext[6]{Image credits: Created by us under the design tutorial of~ \url{http://www.zcool.com.cn/u/1001696}}
\clearpage

\section{Text Effects Transfer between Different Styles, Languages and Fonts}

\footnotetext[7]{Image credits:\\
\textit{water}: \url{http://www.zcool.com.cn/work/ZNTQxOTkzMg==.html}\\
\textit{flame}: \url{http://www.phombo.com/wallpapers/fire-letters-wallpapers-hd-3000x3000-a-z0-9/page-1/}\\
\textit{rust}: Created by us under the design tutorial of \url{http://photo.renren.com/photo/249458089/photo-3396512370/v7}\\
\textit{blink}: Created by us under the design tutorial of \url{http://www.zcool.com.cn/article/ZNTYxNTY=.html}\\
\textit{drop}: Created by us under the design tutorial of \url{http://www.zcool.com.cn/work/ZNDk1MzQxMg==.html}\\
\textit{silver} Created by us under the design tutorial of \url{http://www.zcool.com.cn/work/ZMTc2NzU0OA==.html}
}

\def\srcNameA {4}
\def\srcNameB {11}
\def\srcNameC {16}
\def\srcEffectsA {flame}
\def\srcEffectsB {rust}
\def\srcEffectsC {drop}
\def\trgNameA {21}
\def\trgNameB {23}
\def\trgNameC {24}
\def\trgNameD {25}

\begin{figure}[H]
\centering
\subfigure{
\includegraphics[width=0.22\linewidth]{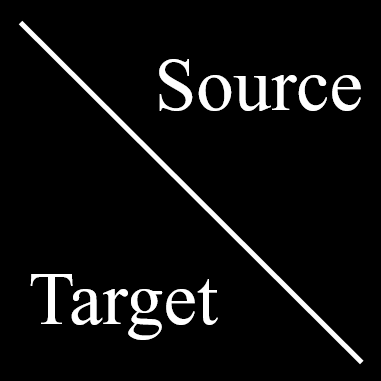}}
\subfigure{
\includegraphics[width=0.22\linewidth]{supp/\srcNameA-e.jpg}}
\subfigure{
\includegraphics[width=0.22\linewidth]{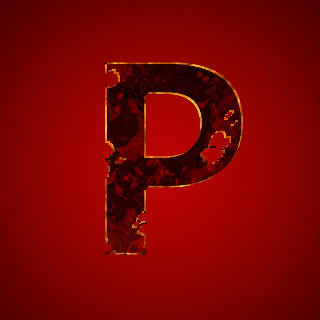}}
\subfigure{
\includegraphics[width=0.22\linewidth]{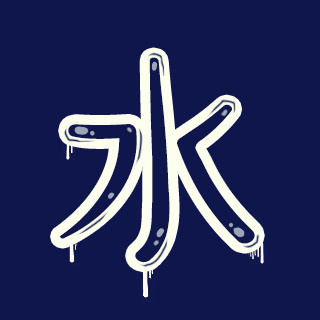}}\vspace{-2mm}
\subfigure{
\includegraphics[width=0.22\linewidth]{supp/\trgNameA-text.jpg}}
\subfigure{
\includegraphics[width=0.22\linewidth]{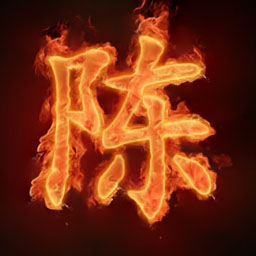}}
\subfigure{
\includegraphics[width=0.22\linewidth]{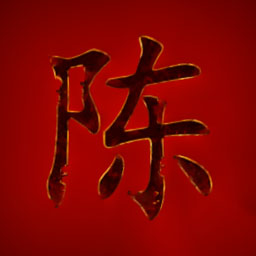}}
\subfigure{
\includegraphics[width=0.22\linewidth]{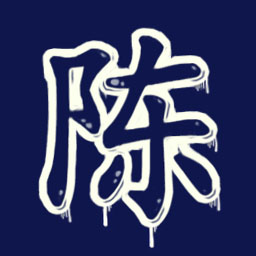}}\vspace{-2mm}
\subfigure{
\includegraphics[width=0.22\linewidth]{supp/\trgNameB-text.jpg}}
\subfigure{
\includegraphics[width=0.22\linewidth]{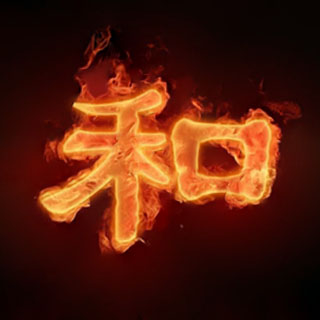}}
\subfigure{
\includegraphics[width=0.22\linewidth]{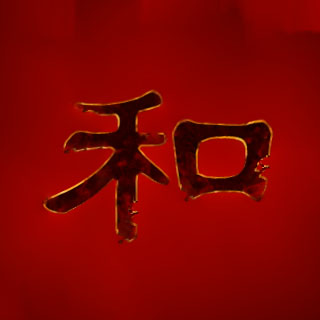}}
\subfigure{
\includegraphics[width=0.22\linewidth]{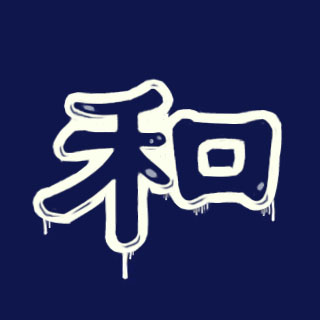}}\vspace{-2mm}
\subfigure{
\includegraphics[width=0.22\linewidth]{supp/\trgNameC-text.jpg}}
\subfigure{
\includegraphics[width=0.22\linewidth]{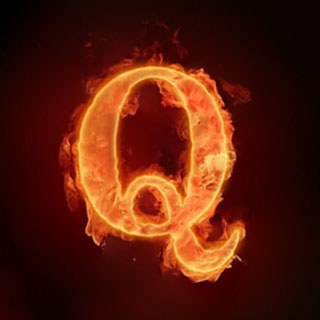}}
\subfigure{
\includegraphics[width=0.22\linewidth]{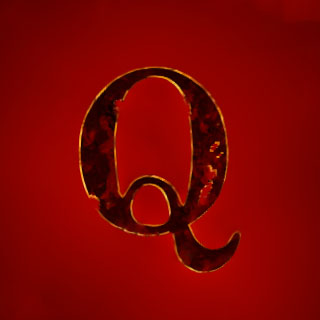}}
\subfigure{
\includegraphics[width=0.22\linewidth]{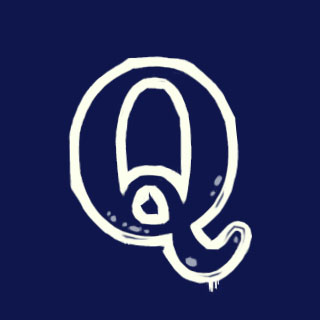}}\vspace{-2mm}
\subfigure{
\includegraphics[width=0.22\linewidth]{supp/\trgNameD-text.jpg}}
\subfigure{
\includegraphics[width=0.22\linewidth]{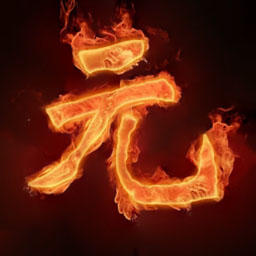}}
\subfigure{
\includegraphics[width=0.22\linewidth]{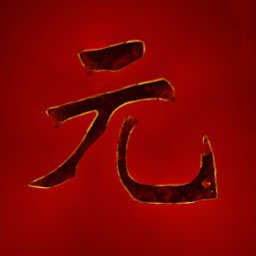}}
\subfigure{
\includegraphics[width=0.22\linewidth]{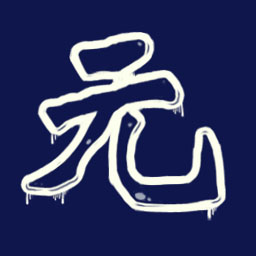}}
\caption{Results of our method with  \textit{\srcEffectsA},  \textit{\srcEffectsB} and \textit{\srcEffectsC} as examples. Top row: three source text effects images. Left column: four target characters.}
\label{fig:resultA}
\end{figure}

\def\trgNameA {26}
\def\trgNameB {27}
\def\trgNameC {29}
\def\trgNameD {30}

\begin{figure}[H]
\centering
\subfigure{
\includegraphics[width=0.22\linewidth]{supp/blank.png}}
\subfigure{
\includegraphics[width=0.22\linewidth]{supp/\srcNameA-e.jpg}}
\subfigure{
\includegraphics[width=0.22\linewidth]{supp/\srcNameB-e.jpg}}
\subfigure{
\includegraphics[width=0.22\linewidth]{supp/\srcNameC-e.jpg}}\vspace{-2mm}
\subfigure{
\includegraphics[width=0.22\linewidth]{supp/\trgNameA-text.jpg}}
\subfigure{
\includegraphics[width=0.22\linewidth]{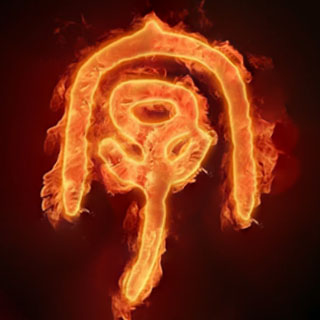}}
\subfigure{
\includegraphics[width=0.22\linewidth]{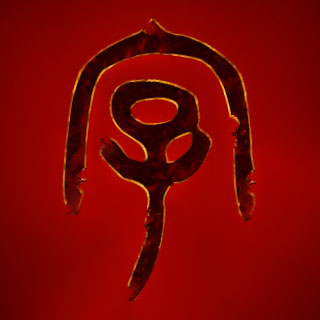}}
\subfigure{
\includegraphics[width=0.22\linewidth]{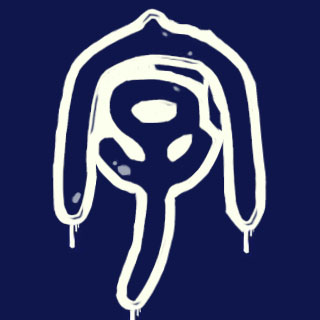}}\vspace{-2mm}
\subfigure{
\includegraphics[width=0.22\linewidth]{supp/\trgNameB-text.jpg}}
\subfigure{
\includegraphics[width=0.22\linewidth]{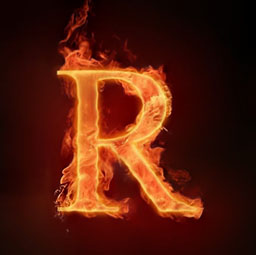}}
\subfigure{
\includegraphics[width=0.22\linewidth]{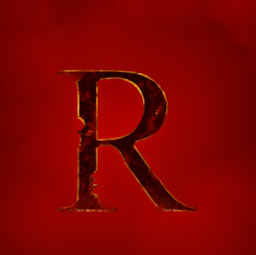}}
\subfigure{
\includegraphics[width=0.22\linewidth]{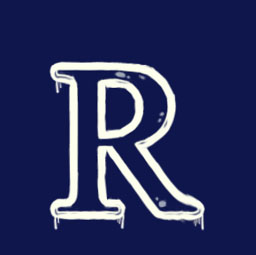}}\vspace{-2mm}
\subfigure{
\includegraphics[width=0.22\linewidth]{supp/\trgNameC-text.jpg}}
\subfigure{
\includegraphics[width=0.22\linewidth]{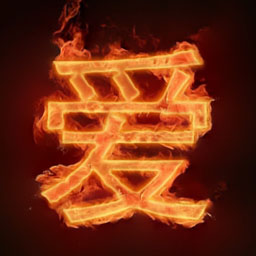}}
\subfigure{
\includegraphics[width=0.22\linewidth]{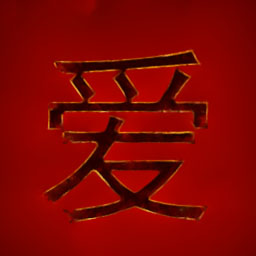}}
\subfigure{
\includegraphics[width=0.22\linewidth]{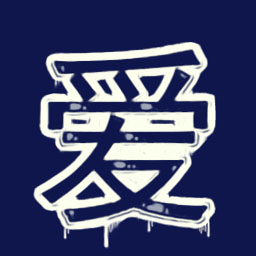}}\vspace{-2mm}
\subfigure{
\includegraphics[width=0.22\linewidth]{supp/\trgNameD-text.jpg}}
\subfigure{
\includegraphics[width=0.22\linewidth]{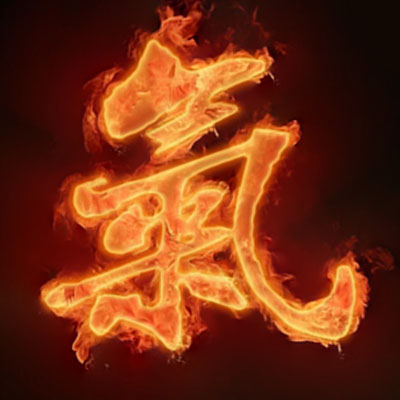}}
\subfigure{
\includegraphics[width=0.22\linewidth]{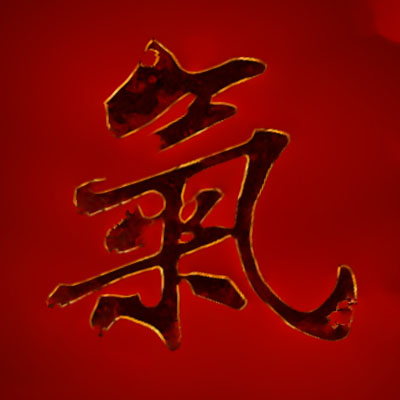}}
\subfigure{
\includegraphics[width=0.22\linewidth]{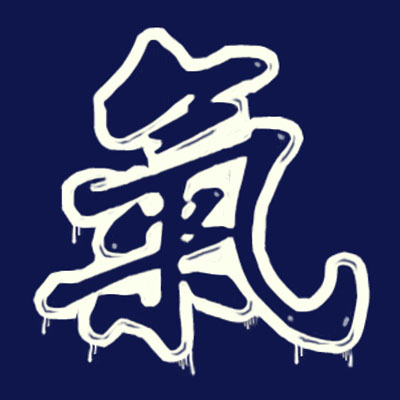}}
\caption{Results of our method with  \textit{\srcEffectsA},  \textit{\srcEffectsB} and \textit{\srcEffectsC} as examples. Top row: three source text effects images. Left column: four target characters.}
\label{fig:resultB}
\end{figure}

\def\srcNameA {1}
\def\srcNameB {14}
\def\srcNameC {19}
\def\srcEffectsA {water}
\def\srcEffectsB {blink}
\def\srcEffectsC {silver}
\def\trgNameA {21}
\def\trgNameB {23}
\def\trgNameC {24}
\def\trgNameD {25}

\begin{figure}[H]
\centering
\subfigure{
\includegraphics[width=0.22\linewidth]{supp/blank.png}}
\subfigure{
\includegraphics[width=0.22\linewidth]{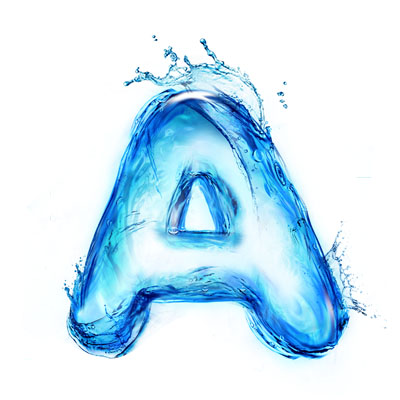}}
\subfigure{
\includegraphics[width=0.22\linewidth]{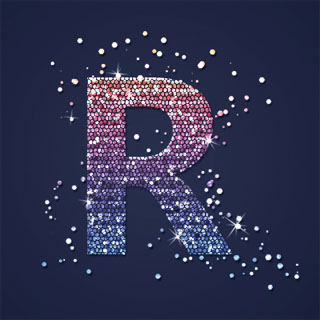}}
\subfigure{
\includegraphics[width=0.22\linewidth]{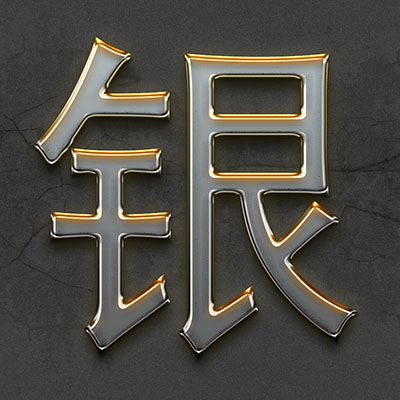}}\vspace{-2mm}
\subfigure{
\includegraphics[width=0.22\linewidth]{supp/\trgNameA-text.jpg}}
\subfigure{
\includegraphics[width=0.22\linewidth]{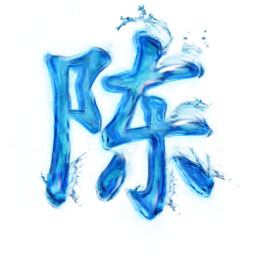}}
\subfigure{
\includegraphics[width=0.22\linewidth]{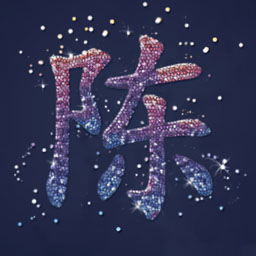}}
\subfigure{
\includegraphics[width=0.22\linewidth]{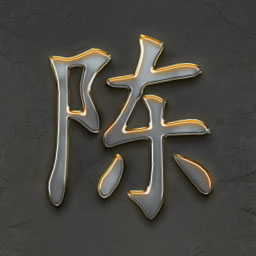}}\vspace{-2mm}
\subfigure{
\includegraphics[width=0.22\linewidth]{supp/\trgNameB-text.jpg}}
\subfigure{
\includegraphics[width=0.22\linewidth]{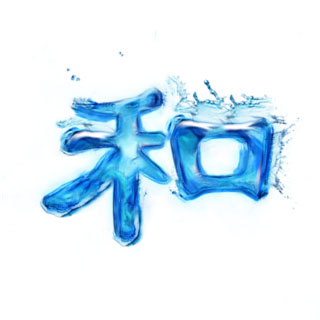}}
\subfigure{
\includegraphics[width=0.22\linewidth]{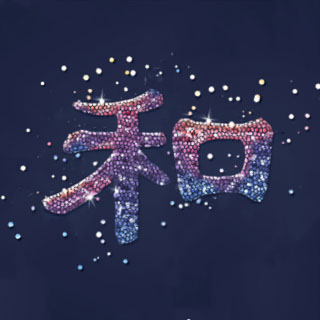}}
\subfigure{
\includegraphics[width=0.22\linewidth]{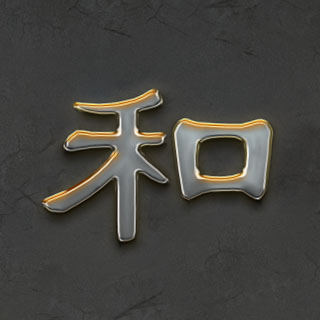}}\vspace{-2mm}
\subfigure{
\includegraphics[width=0.22\linewidth]{supp/\trgNameC-text.jpg}}
\subfigure{
\includegraphics[width=0.22\linewidth]{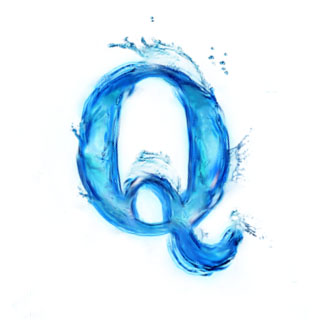}}
\subfigure{
\includegraphics[width=0.22\linewidth]{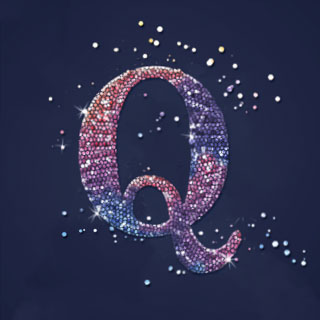}}
\subfigure{
\includegraphics[width=0.22\linewidth]{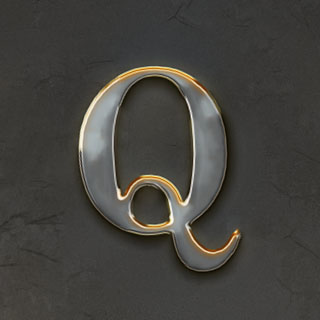}}\vspace{-2mm}
\subfigure{
\includegraphics[width=0.22\linewidth]{supp/\trgNameD-text.jpg}}
\subfigure{
\includegraphics[width=0.22\linewidth]{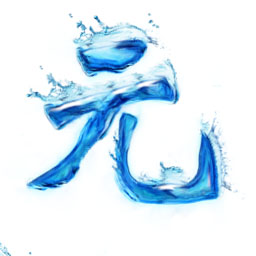}}
\subfigure{
\includegraphics[width=0.22\linewidth]{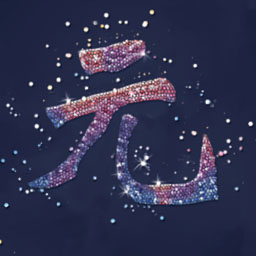}}
\subfigure{
\includegraphics[width=0.22\linewidth]{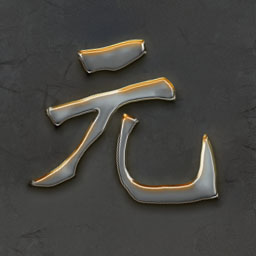}}
\caption{Results of our method with  \textit{\srcEffectsA},  \textit{\srcEffectsB} and \textit{\srcEffectsC} as examples. Top row: three source text effects images. Left column: four target characters.}
\label{fig:resultC}
\end{figure}

\def\trgNameA {26}
\def\trgNameB {27}
\def\trgNameC {29}
\def\trgNameD {30}

\begin{figure}[H]
\centering
\subfigure{
\includegraphics[width=0.22\linewidth]{supp/blank.png}}
\subfigure{
\includegraphics[width=0.22\linewidth]{supp/\srcNameA-e.jpg}}
\subfigure{
\includegraphics[width=0.22\linewidth]{supp/\srcNameB-e.jpg}}
\subfigure{
\includegraphics[width=0.22\linewidth]{supp/\srcNameC-e.jpg}}\vspace{-2mm}
\subfigure{
\includegraphics[width=0.22\linewidth]{supp/\trgNameA-text.jpg}}
\subfigure{
\includegraphics[width=0.22\linewidth]{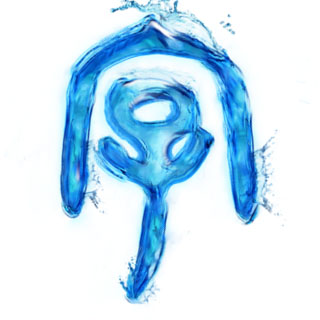}}
\subfigure{
\includegraphics[width=0.22\linewidth]{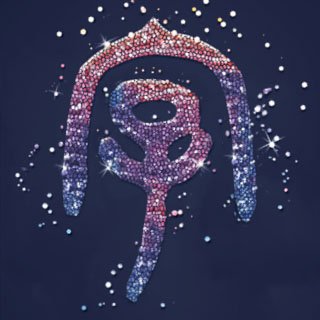}}
\subfigure{
\includegraphics[width=0.22\linewidth]{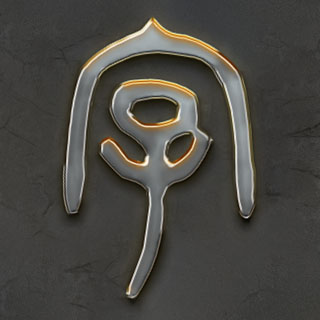}}\vspace{-2mm}
\subfigure{
\includegraphics[width=0.22\linewidth]{supp/\trgNameB-text.jpg}}
\subfigure{
\includegraphics[width=0.22\linewidth]{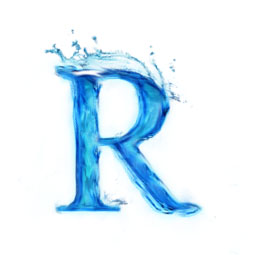}}
\subfigure{
\includegraphics[width=0.22\linewidth]{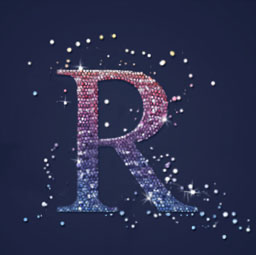}}
\subfigure{
\includegraphics[width=0.22\linewidth]{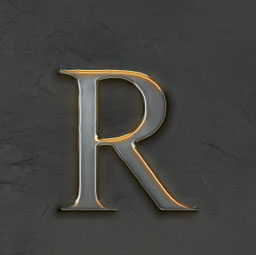}}\vspace{-2mm}
\subfigure{
\includegraphics[width=0.22\linewidth]{supp/\trgNameC-text.jpg}}
\subfigure{
\includegraphics[width=0.22\linewidth]{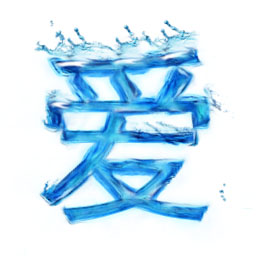}}
\subfigure{
\includegraphics[width=0.22\linewidth]{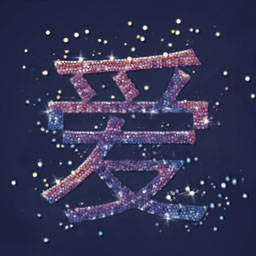}}
\subfigure{
\includegraphics[width=0.22\linewidth]{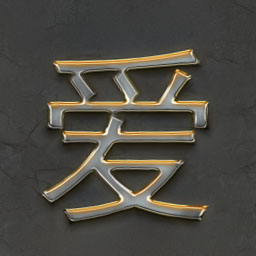}}\vspace{-2mm}
\subfigure{
\includegraphics[width=0.22\linewidth]{supp/\trgNameD-text.jpg}}
\subfigure{
\includegraphics[width=0.22\linewidth]{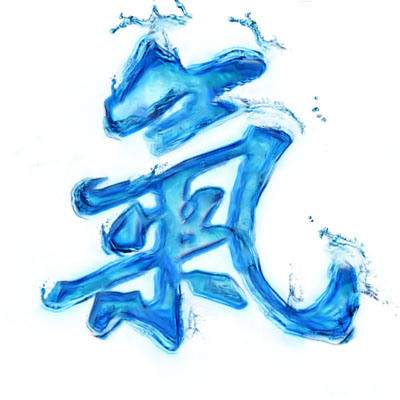}}
\subfigure{
\includegraphics[width=0.22\linewidth]{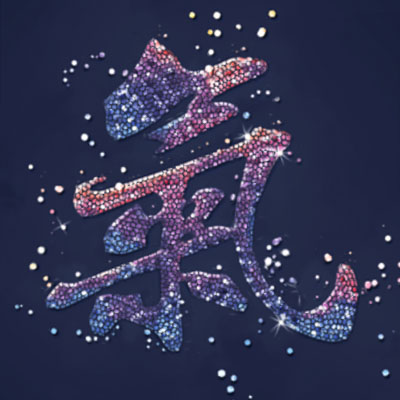}}
\subfigure{
\includegraphics[width=0.22\linewidth]{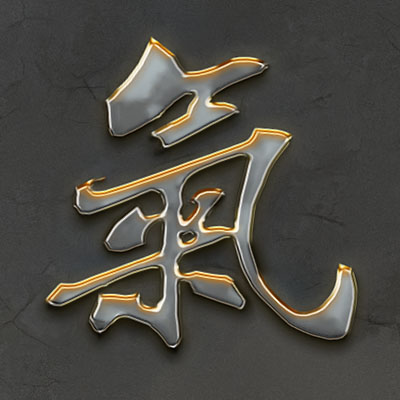}}
\caption{Results of our method with  \textit{\srcEffectsA},  \textit{\srcEffectsB} and \textit{\srcEffectsC} as examples. Top row: three source text effects images. Left column: four target characters.}
\label{fig:resultD}
\end{figure}

\newpage

\section{Artistic Typography Library}

\end{document}